\documentclass[lettersize,journal]{IEEEtran}
\usepackage{amsmath,amsfonts}
\usepackage{algorithmic}
\usepackage{array}
\usepackage[caption=false,font=normalsize,labelfont=sf,textfont=sf]{subfig}
\usepackage{textcomp}
\usepackage{stfloats}
\usepackage{url}
\usepackage{verbatim}
\usepackage{graphicx}
\def\BibTeX{{\rm B\kern-.05em{\sc i\kern-.025em b}\kern-.08em
    T\kern-.1667em\lower.7ex\hbox{E}\kern-.125emX}}
\usepackage{balance}
\usepackage{xcolor}
\usepackage{adjustbox}
\usepackage{flushend}
\usepackage{rotating}
\usepackage[pdfstartview=XYZ,
bookmarks=true,
colorlinks=true,
linkcolor=blue,
urlcolor=blue,
citecolor=blue,
pdftex,
bookmarks=true,
linktocpage=true, % makes the page number as hyperlink in table of content
hyperindex=true
]{hyperref}

\begin{document}
\title{Incremental procedural and sensorimotor learning in cognitive humanoid robots}

%\subsection{Author Names and Affiliations}
%\noindent The author section should be coded as follows:
%\begin{verbatim}
%\author{Masahito Hayashi 
%\IEEEmembership{Fellow, IEEE}, Masaki Owari
%\thanks{M. Hayashi is with Graduate School 
%of Mathematics, Nagoya University, Nagoya, 
%Japan}
%\thanks{M. Owari is with the Faculty of 
%Informatics, Shizuoka University, 
%Hamamatsu, Shizuoka, Japan.}
%}
%\end{verbatim}

\author{Leonardo de Lellis Rossi, Letícia Mara Berto, Eric Rohmer, Paula Paro Costa, \\Ricardo Ribeiro Gudwin,  Esther Luna Colombini and Alexandre da Silva Simões
\thanks{This work was developed within the scope of PPI-Softex with support from MCTI through the Technical Cooperation Term [01245.013778/2020-21].}
\thanks{A. S. Simões and L. L. Rossi are with Dept. of Control and Automation Engineering (DECA), Institute of Science and Technology (ICT), Campus Sorocaba, Universidade Estadual Paulista (Unesp)}
\thanks{R. R. Gudwin, E. Rohmer, P. P. Costa and L. L. Rossi are with the Dept. of Computer Engineering and Industrial Automation (DCA), School of Electrical and Computer Engineering (FEEC), University of Campinas, Brazil}
\thanks{E. L. Colombini and L. M. Berto are with the Laboratory of Robotics and Cognitive Systems (LaRoCS), Institute of Computing, University of Campinas, Brazil}
\thanks{All authors are with the Artificial Intelligence and Cognitive Architectures Hub (H.IAAC), University of Campinas, Brazil}}

\markboth{Preprint submitted to IEEE Transactions on Cognitive and Developmental Systems}%
{Incremental procedural and sensorimotor learning in cognitive humanoid robots}

\maketitle

\begin{abstract}
The ability to automatically learn movements and behaviors of increasing complexity is a long-term goal in autonomous systems. Indeed, this is a very complex problem that involves understanding how knowledge is acquired and reused by humans as well as proposing mechanisms that allow artificial agents to reuse previous knowledge. Inspired by Jean Piaget's theory's first three sensorimotor substages, this work presents a cognitive agent based on CONAIM (\emph{Conscious Attention-Based Integrated Model}) that can learn procedures incrementally. Throughout the paper, we show the cognitive functions required in each substage and how adding new functions helps address tasks previously unsolved by the agent. Experiments were conducted with a humanoid robot in a simulated environment modeled with the  Cognitive Systems Toolkit (CST) performing an object tracking task. The system is modeled using a single procedural learning mechanism based on Reinforcement Learning. The increasing agent's cognitive complexity is managed by adding new terms to the reward function for each learning phase. Results show that this approach is capable of solving complex tasks incrementally. 

%Incrementally learning to solve problems with increasing complexity by reusing previous skills is a long-term goal in autonomous systems. Indeed, this is a very complex problem that involves understanding how knowledge is acquired and reused. One step towards this goal involves finding mechanisms that, motivated by the agent's learning maturity, aim towards more complex goals while reusing previous knowledge. This work presents a cognitive architecture based on the CONAIM model (\emph{Conscious Attention-Based Integrated Model}) that can learn procedures incrementally. The agent was evaluated with experiments inspired by the first three sensorimotor substages of Jean Piaget's Theory. Throughout the paper, we show the set of cognitive functions required in each substage and how adding new functions helps address tasks previously unsolved by the agent. We build a single procedural learning mechanism based on Reinforcement Learning, adding terms to the reward function for each phase. The implementation was performed with the \emph{Cognitive Systems Toolkit} (CST) and validated with mobile robots in simulated environments. Results shown that this approach can solve complex object tracking experiments incrementally. 

\end{abstract}

\begin{IEEEkeywords}
Cognitive Robotics, Cognitive Architectures, Reinforcement Learning, Incremental Learning, Developmental Robotics.
\end{IEEEkeywords}

\section{Introduction}

\IEEEPARstart{A}{dvancements} in artificial intelligence and robotics increased the interest in introducing robots into daily activities that involve interaction with other agents, both robots and humans. These robots should operate autonomously in complex, partially unknown, unpredictable, and unstructured scenarios, making pre-programming impossible and requiring robots to have a superior capability to perform tasks. This challenge raises questions such as how to incorporate new knowledge and skills through interactions with the world, resulting in the research area of Cognitive Robotics. Cognitive Robotics is intrinsically related to Cognitive Architectures (CA), which represent comprehensive computer models providing theoretical frameworks to work with cognitive processes searching for complex behavior. 

Cognitive architectures are systems that can reason in different domains, develop different views, adapt to new situations, and reflect on themselves \cite{reggia2013, simoes_2015}. They are general control systems inspired by scientific theories developed to explain cognition in humans and other animals, comprising modules responsible for implementing different cognitive abilities, such as perception, attention, memory, reasoning, and learning.

Inspired by how humans build knowledge through interactions with the world, cognitive architecture researchers seek to reproduce this behavior with artificial creatures \cite{paraense2016_2}. However, the development of cognitive skills in machines requires the coordination of complex mechanisms that depend on each other. According to Piaget \cite{piaget_origins_1952}, the process of developing these skills is incremental and evolutionary.

In this work, a cognitive agent based on the CONAIM model (\emph{Conscious Attention-Based Integrated Model}) \cite{simoes_2015} was proposed and implemented with the \emph{Cognitive Systems Toolkit} \cite{paraense2016_2}. A humanoid robot was designed to incrementally learn procedures to perform object tracking experiments inspired by the first three sensorimotor substages of Jean Piaget's Theory \cite{piaget_origins_1952}. Throughout the work, we present the cognitive functions necessary to form circular reactions in each substage using a Reinforcement Learning (RL) \cite{sutton1998} environment and how new functions can be added to the reward function allowing the agent to solve complex tasks, previously unresolved. %demonstrating that the agent can learn new behaviors based on the reuse of previous knowledge.

As the main contributions of this work, we can list the following:

\begin{enumerate}
    \item The proposition of a cognitive architecture based on CONAIM with attention, memories, and learning modules focused on sensorimotor and procedural learning; 
    \item The design and implementation of CONAIM's top-down pathway in CST that can be incorporated into any agent implemented with CST;
    \item The design and implementation of a single procedural learning mechanism in CST that can incrementally learn and reuse schemas for the first three sensorimotor substages of Piaget's Theory;
    \item The modeling of a set of environments for sensorimotor experiments for the movements learning in humanoid robots;
     \item The implementation and evaluation of sensorimotor experiments for object tracking in the first three sensorimotor substages of Piaget's Theory as proposed by \cite{berto_2020_thesis}.
\end{enumerate}

%The relationship between attention, learning and position at each instant of physical and virtual actuators was also investigated, through the analysis of the relationship of cognitive development with memories and learning in the CONAIM model. 
The code used to implement the architecture is available at: \textbf{\url{https://github.com/CST-Group/cst}}.

\section{Cognitive Architectures}

\noindent Cognitive architectures are systems that can reason in different domains, develop different views, adapt to new situations, and reflect on themselves \cite{kotseruba2020}. They are general control systems inspired by the cognition of humans and other animals, comprising modules responsible for implementing different cognitive abilities, such as perception, attention, memory, reasoning, and learning \cite{paraense2016_2}.

A cognitive architecture plays the role of cognition in computational modeling, making explicit the set of processes and assumptions on which this cognitive model is based \cite{vernon_desiderata_2016}. It consists of processing units that represent, extract, select and combine knowledge and memories to produce behavior \cite{anderson1998, laird_soar_2008, thomson2014}.

Next, the main aspects of the reference cognitive architecture used in this work and the computational toolkit used to program the cognitive agent are described. Both were proposed in previous work by the working group.

\subsection{CONAIM}
\noindent The CONAIM model (\emph{Conscious Attention-Based Integrated Model}) \cite{simoes_2015, simoes_2017, colombini2017} is a formal attention-based model for machine consciousness. CONAIM incorporates several relevant aspects for a cognitive agent (memories, body schema, motivation, attention, among others) and is capable of dealing with multiple sensory systems, multiple processes of feature extraction, decision making, and learning \cite{colombini2017}.
The model provides a consciousness-based agent that performs calculations on attention-directed schemas, significantly reducing the space of the model's input dimensions.
In its cognitive cycle, \textbf{top-down} and \textbf{bottom-up} mechanisms are used. During the \textbf{bottom-up} cycle, the sensors provide external data. The data is stored in the sensory memory to form feature, attentional and salience maps \cite{colombini2017}. The complete modeling is detailed in \cite{colombini2017}.
%These maps are combined into a single combined feature map, which represents a weighted sum of features. The combination of feature maps with the attentional map previously generated by the agent generates a map of saliences, a representation of the environment perceived by the agent, modulated by the attentional process. Through a selection of winners, for example by a \emph{winner-takes-all} (WTA) algorithm, the most relevant stimulus can be defined in the saliency map. The model also considers a feedback inhibition (IOR) mechanism. The IOR promotes the decay of the sensitivity of a region when there is a prolonged display of the stimulus. Thus, promoting attention to the next location with the second highest value on the saliency map. The complete modeling is detailed in \cite{colombini2017}.
In the \textbf{top-down} cycle, the attentional modulation of the system will depend on the global state of attention (for example, the robot's battery level). It will also depend on the agent's objective, and the attentional dynamics of the current state \cite{colombini2014}. 

\subsection{CST}
\noindent The \emph{Cognitive Systems Toolkit} \cite{paraense2016_2} is an open-source Java-based \cite{cstGIT} toolkit for building cognitive architectures.
%The CST allows the use and integration of several computational technologies found in other architectures, such as CLARION, SOAR and LIDA. 
The core of CST consists of a set of basic concepts that can be generalized within any cognitive architecture built.  CST tools allow the creation of multi-agent systems running asynchronously and in parallel. 
%Your cognitive functions are classes that can be combined into a parallel set of interaction devices. CST tools allow the creation of multi-agent systems that run asynchronously and in parallel.
The CST architecture is \textbf{codelet} oriented. Codelets are small pieces of code, implemented as asynchronous functions, and that run in parallel with simple and defined tasks \cite{paraense2016_2}.
%Codelets represent all major cognitive functions, occurring constantly and cyclically, and are responsible for system behaviors.
%Each codelet in the CST has two entries, Local (Li) and Global (Gi) \cite{paraense2016_2}. The \textbf{Local (Li)} entry is the default entry for information about selected memory objects and is usually fixed but can be changed through learning mechanisms. The \textbf{Global (Gi)} entry is used to get information from the \emph{Global Workspace} (GWT) and is updated all the time due to attention mechanisms. The \emph{codelets} in the CST also have two outputs, Default (O) and Activation Level (A). The \textbf{Default (O)} output is used to change or create new information in memory. The output \textbf{Activation level (A)} indicates the relevance of the information obtained in the Standard output (O) and is used to select the information that will be used in the GWT.
A \textbf{memory object} (MO) is a signal or representation used, with other MOs, by codelets to store and access data \cite{paraense2016_2}.
%The main property of an MO is its information (I), modeled as a generic object in Java, which can be represented as a unitary value in lists or collections. Each MO also has a time stamp (T), which indicates its last update, and an evaluation (E).

\section{Developmental Robotics}
\noindent The development of artificial agents with autonomy, adaptive behavior and incremental learning capabilities are research goals in Cognitive Robotics and Developmental Robotics (DevRobotics) \cite{beer_dynamics_2003, cangelosi2014}.
%As biological agents provide the best examples of such behavior in the real world, they are the source of concepts and design principles for artificial agents \cite{ bellas_cognitive_2010}. One such principle is that biological cognition can be described as fundamentally concerned with the manipulation and utilization of memory, perception, thought, and action.
The area emerged due to the need for robots to perform tasks that require comparable levels of human intelligence in complex and unpredictable environments involving adaptation and evolution \cite{stoytchev2009}. The models and experiments in the area are inspired by the principles and mechanisms of development observed in early life, involving robots performing the same cognitive abilities as children, as in the experiments proposed by Jean Piaget.

\subsection{Piaget's Theory}
\noindent A relevant concept in Piaget's theory \cite{piaget_origins_1952} are the \textbf{schemas}, which represent networks of mental structures that help remember specific concepts and understand the environment. When simple mental processes become more sophisticated, new schemas are developed, and behavior becomes more complex and suited to the environment \cite{piaget2008}.

In mental development, according to Piaget, \textbf{adaptation} -- or learning -- is the tendency to adjust mental processes to the environment by changing cognitive structures \cite{piaget_origins_1952}. The adaptation process involves balancing the processes of assimilation and accommodation. Assimilation and Accommodation are inseparable, complementary and simultaneous processes \cite{piaget_origins_1952}. \textbf{Assimilation} is the creation of new schemas following the same cycle or sequence of existing schemas for interpreting experiences and making decisions. \textbf{Accommodation} is the complementary process that involves altering existing schemas as a result of new information acquired through assimilation.
%In this case, a subject will try to solve a given situation by modifying the old structures and looking for new ways of acting to master new scenarios \cite{piaget2008,woolfolk2003}.
Assimilation can originate \textbf{circular reactions}, repetitions of cycles acquired or in the process of acquisition \cite{piaget_origins_1952, piaget_1971}. The circular reaction results from the assimilation of an interesting result unknown to the subject, which was produced by the rediscovery or repetition of the action.
% Circular reactions are processes by which an individual learns to reproduce desired occurrences originally discovered by chance and which brought about a satisfactory result.
Circular reactions can be primary, secondary, or tertiary:
\begin{itemize}
    \item \textbf{Primary Circular Reactions} are behaviors derived from reflexes, activities of the body itself that form new schemes through the coordination of the senses;
\item \textbf{Secondary Circular Reactions} are derived from intentional behaviors that direct interest to external outcomes rather than the baby's body;
\item \textbf{Tertiary Circular Reactions} are the subject's effort to seek new experiences.
\end{itemize}

\subsection{Sensorimotor Experiments for Incremental Learning}

\noindent Experiments with cognitive architectures in the field of DevRobotics are heavily based on theories of childhood development. However, a common point in these experiments is the lack of standardization to conduct and evaluate agent development. To help resolve these issues, we proposed in previous work \cite{berto_2020_thesis} a set of incremental experiments for the robotic scenario according to Piaget's sensorimotor stages of development, along with the expected results. These experiments are based on Piaget's studies \cite{piaget_origins_1952} on the sensorimotor period, on the Bayley Child Development Scale \cite{Michalec2011}, in different scenarios described in the literature for the assessment of learning development in infants \cite{boyd2011, papalia2013, guerin_survey_2013}, and in the parameters and levels of ConsScale \cite{consscaleSite}. The computational scenarios focused on incorporating human behaviors in robots and evaluating their cognitive development. Experiments are classified according to the type of skill to be learned by the agent in many scenarios. 

%She also defined the relationship between \emph{cognits} \cite{fuster2006} and the schemes proposed by Piaget in the development of the sensory-motor stage \cite{piaget_origins_1952}. The concept of \emph{cognit} \cite{fuster2006} is used to represent memory processing elements with an evaluation level \cite{berto_2020_thesis}. The \emph{cognits} were inserted in the context of Piaget's sensorimotor stage, in the following characterization:
%\begin{itemize}
%    \item \textbf{1st Substage. Use of Reflexes:} creation of innate \emph{cognits} (actions, states and learning mechanism);
%\item \textbf{2nd Substage. Primary Circular Reactions:} creation and adaptation of \emph{cognits} based on functions, but not on intentions;
%\item \textbf{3rd Substage. Secondary Circular Reactions:} creation and adaptation of \emph{cognits} to form plans, goals, and intentions. 
%\end{itemize}

\section{Proposed Approach}
\label{sec:proposed_approach}

In the current work, we addressed a subset of the theoretical scenarios proposed by \cite{berto_2020_thesis}, focusing on building an agent to perform exclusively those experiments related to \textbf{object tracking} task, as shown in Table \ref{tab:track}. In those experiments, we expect a robot to incrementally learn the skills of tracking objects in a scene using RGB-D sensors. The goal is to investigate incremental computational processes that can allow robots to learn intentional sensorimotor schemas from an initially unintentional perspective, that is, computational processes able to model circular reactions using procedural memory elements.

%\begin{figure}[ht!]
%    \centering
%    \includegraphics[width=0.49\textwidth]{img/track.jpg}
%    \caption{Subset of experiments related to learning how to track an object in a scene. Adapted from \cite{berto_2020_thesis}.}
%    \label{tab:track}
%\end{figure}

\begin{table*}[t]
%    \begin{adjustbox}{max width=0.48\textwidth}
    %\begin{large}
    \centering
    \begin{tabular}{|p{1cm}|p{4cm}|p{4cm}|p{7.0cm}|}
    \hline
    \multicolumn{4}{|c|}{\textbf{Task: Tracking objects using a vision sensor}} \\ \hline
    \textbf{Sub-stage} & \textbf{Experiment} & \textbf{Expected results} & \textbf{Details} \\  \hline
    1 & Robot placed in front of an object at close distance & Staring at the object & Object properties: \begin{itemize}
        \item \textbf{position: } inside the agent's visual field
        \item \textbf{movement: }  fixed
        \item \textbf{color: }primary 
    \end{itemize}
    Agent's visual acuity: 20/400 Agent's FOV: small \\  \hline
    1 & Robot placed in front of an object at close distance & Track object with the look & Object properties: \begin{itemize}
        \item \textbf{position: } inside the agent's visual field
        \item \textbf{movement: }  moving slowly
        \item \textbf{color: }primary 
    \end{itemize}
    Agent's visual acuity: 20/400 Agent's FOV: small \\ \hline
    1 & Robot placed in front of an object  at close distance & Do not follow the object with the look & Object properties: \begin{itemize}
        \item \textbf{position: } leaving the agent's visual field
        \item \textbf{movement: }  moving slowly
        \item \textbf{color: }primary 
    \end{itemize}
    Agent's visual acuity: 20/400 Agent's FOV: small \\ \hline
    2 & Robot placed in front of an object at medium distance & Track object with the look &  Object properties: \begin{itemize}
        \item \textbf{position: } inside the agent's visual field
        \item \textbf{movement: }  moving at medium speed
        \item \textbf{color: }primary 
    \end{itemize}
    Agent's visual acuity: 20/100 Agent's FOV: increased FOV \\ \hline
    2 & Robot placed in front of an object at medium distance & Track object with the look &  Object properties: \begin{itemize}
        \item \textbf{position: } outside the agent's visual field
        \item \textbf{movement: }  fixed position
        \item \textbf{color: }primary 
    \end{itemize}
    Agent's visual acuity: 20/100 Agent's FOV: increased FOV \\  \hline
     2 & Robot placed in front of an object at medium distance & Accompany object while looking within visual field. Does not follow outside it & 
      Object properties: \begin{itemize}
        \item \textbf{position: } entering or leaving the agent's visual field
        \item \textbf{movement: }  medium speed
        \item \textbf{color: }primary 
    \end{itemize}
    Agent's visual acuity: 20/100 Agent's FOV: increased FOV \\  \hline
    3 &  Robot placed in front of an object at high distance & Track object with the look & Object properties: \begin{itemize}
        \item \textbf{position: } inside the visual field
        \item \textbf{movement: }  medium speed
        \item \textbf{color: }primary 
    \end{itemize}
    Agent's visual acuity: 20/20 Agent's FOV: increased FOV \\  \hline
    3 & Robot placed in front of an object at high distance & Accompany object with the look (turns the head to continue accompanying the object) & Object properties: \begin{itemize}
        \item \textbf{position: } leaving or entering the visual field
        \item \textbf{movement: }  medium speed
        \item \textbf{color: }primary 
    \end{itemize}
    Agent's visual acuity: 20/20 Agent's FOV: increased FOV \\  \hline
    \end{tabular}
  %  \end{large}
 %   \end{adjustbox}
    \newline
    \caption{Subset of experiments and expected results for a cognitive robot sensorimotor learning in a visual object tracking task. Adapted from \cite{berto_2020_thesis}.}
    \label{tab:track}
\end{table*}

The experiments proposed for object tracking describe what capabilities are expected from the agent in this task at each developmental stage and what it should not accomplish. These experiments are designed for scenarios with increasing complexity. It also defines which sensors should be employed to achieve such capabilities and any sensor limitations. For example, it considers the child's visual acuity development when vision is employed. Objects are positioned at different distances from the robot but are perceived according to the visual acuity compatible with the presented by a baby at the specified sensorimotor substage. We can expect a less precise associated behavior when considering a less accurate sensor, as discussed later in the experiments. The agent's visual acuity and its distance to the objects are variable in the experiments. %The change in visual acuity is simulated by changing the camera resolution along substages, and the agent's behavior refining along substages is simulated by adding new actions to the robot action list. 
We assess the development level of the robot by verifying if it can achieve the expected result, as described in Table \ref{tab:track}. %Experiments were performed with increasing difficulty, reusing previously learned schemas to achieve more complex behaviors in more challenging scenarios. 

\section{Methodology}

%\noindent This section describes several aspects of the investigation performed.

%A cognitive robot model was designed to investigate which cognitive modules should be present in an agent that can learn incrementally to perform cognitive experiments with increasing levels of difficulty. 

%A humanoid robot was modeled using CONAIM modules implemented in CST. Modules were added as the agent evolves from the 1st to the 3rd sensorimotor substage, so that it responds to more complex tasks. 
%The active modules for each substage are shown in Figure \ref{fig:model}.
%Reinforcement learning was employed for the agent to learn schemas for procedural memory. To match the object tracking experiments \cite{berto_2020_thesis}, the robot's visual acuity was increased at each substage by changing the camera's resolution. The list of available behaviors was also increased, refining the actions as the agent grows.

\subsection{Robots and Environment}

\begin{figure*}[ht!]
\begin{adjustbox}{max width=\textwidth}
\centering
\begin{tabular}{cccc}
  \includegraphics[width=0.225\textwidth]{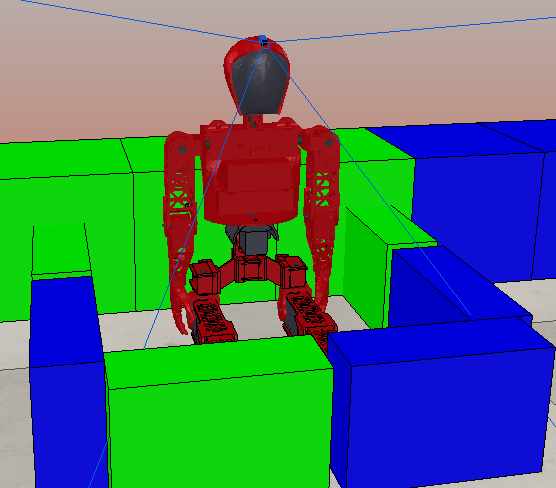}
   &\includegraphics[width=0.235\textwidth]{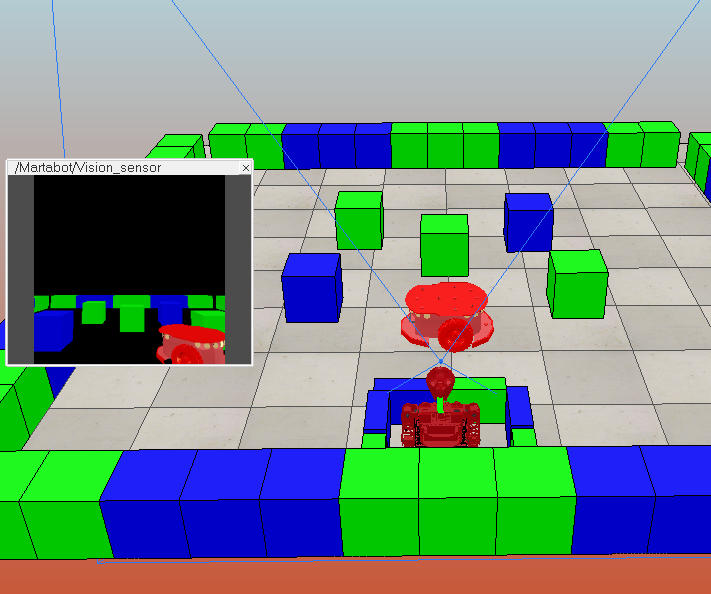} 
    &\includegraphics[width=0.225\textwidth]{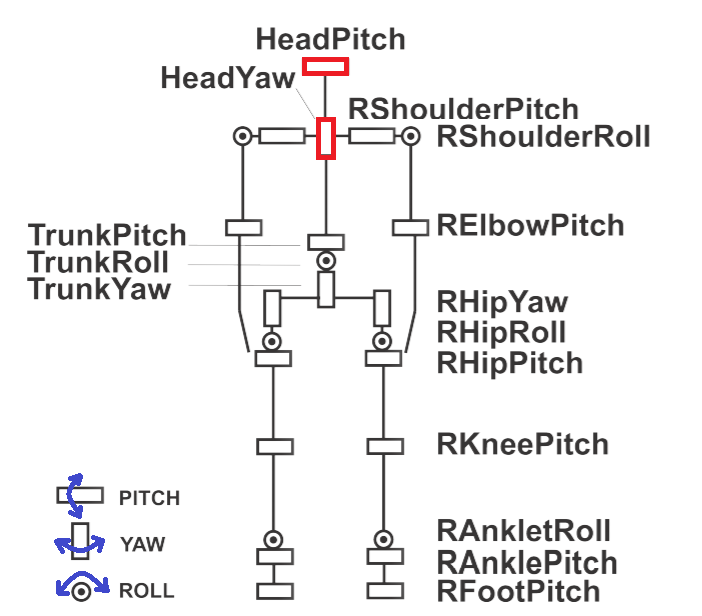} 
   &\includegraphics[width=0.214\textwidth]{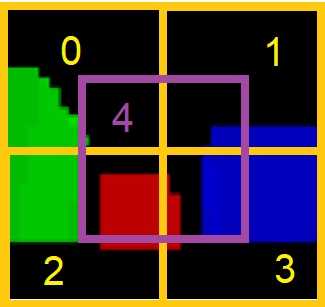} 
    \\
    (a) & (b) & (c) & (d)\\ 
\end{tabular}   
\end{adjustbox}
\caption{Simulation environment. From Left to Right. a) Marta robot equipped with an RGB-D camera; b) Environment with distributed colored blocks and a Pioneer P3DX robot acting as a distractor. Marta's view of the scene is shown at left. (c) Agent's degrees of freedom. Motors used in red. (d) Division of the agent's visual space for the virtual actuator (named as "eye", "look" or "fovea").}
\label{fig:Marta}
\end{figure*}

\noindent The humanoid robot Marta was adopted in our experiments. Marta is a teen-size 1.1m tall female robot designed and built by our workgroup. Marta has 25 degrees of freedom, and it's head -- particularly relevant in the present work -- can perform \emph{pitch} and \emph{yaw} movements. The robot was equipped with an RGB-D camera on its head, inspecting the world in four distinct channels of color (R, G and B) and distance information (D).
%The CAD model of the robot, weights and dynamic properties were transferred to the \emph{CoppeliaSim} \cite{coppeliaSim} simulator and validated to match the real and simulated dynamic model.
The robot was controlled by a cognitive system detailed in the next sections. Several simulated scenes were also created in CoppeliaSim for the experiments. In the scenes, Marta is sitting in a small space with a wide view of its surroundings. An arena was delimited outside this first space, and colored blocks (blue and green) were randomly distributed. A second robot, a red Pioneer P3DX, randomly navigates the arena as a mobile distractor. This robot was modeled with reactive behaviors using the Braitenberg Algorithm \cite{braitenberg1984}. Both robots and the environment are shown in Figure \ref{fig:Marta}.

\subsection{Robot control overview}

Table \ref{tab:sumario} summarizes the key aspects of the humanoid robot control strategies adopted in the proposed approach, focusing on Piaget's schemas created or utilized along developmental sensorimotor stages. It emphasizes the cognitive system functions present, the configuration of the robot's sensors and actuators, and the Reinforcement Learning reward strategy adopted. We will address these aspects in the following sections.

\begin{table*}[ht!]
    \begin{center}
    \begin{footnotesize}
    \begin{tabular}{ | m{1cm} | m{4cm} |  m{1.5cm} | m{1.2cm}| m{2cm} | m{5.5cm} |} 
        \hline
        Substage & Circular Piaget's schemas & Cognitive system functions & Vision resolution & Actions and resolution & RL Reward strategy \\ 
        \hline
        1 & Use of Reflexes: creation of the first schemas by exercising reflexive behaviors & Standard & Low & 10 (motor and virtual), low resolution &  $\bullet$ Stimulate new procedural memories \newline $\bullet$ Stimulate synchronization between movement and bottom-up attention focus \newline $\bullet$ Punish failures \\ 
        \hline
        2 &  Primary Circular Reactions: creation and adaptation of schemas based on functions, but not intentions & Standard + motivation & Medium & 10 (motor and virtual), high resolution & $\bullet$ Stimulate new procedural memories \newline $\bullet$ Stimulate synchronization between movement and bottom-up attention focus \newline $\bullet$ Punish failures\\ 
        \hline
        3 & Secondary Circular Reactions: creation and adaptation of schemas to form plans, objectives, and intentions & Standard + motivation & High & 17 (motor, virtual, attentional), high resolution & $\bullet$ Stimulate new procedural memories \newline $\bullet$ Stimulate synchronization between movement and bottom-up attention focus \newline $\bullet$ Stimulate synchronization between movement and top-down attention (intention) \newline $\bullet$ Punish failures\\
        \hline
    \end{tabular}
    \end{footnotesize}
    \end{center}
    \caption{Overview of the approach: schemas, vision resolution, actions and resolutions, cognitive system and reward strategy adopted in each substage in the experiments.}
    \label{tab:sumario}
\end{table*}

\subsection{Cognitive model}

\begin{figure*}[hbt!]
    \centering
    
    \begin{adjustbox}{max width=\textwidth}
    \begin{tabular}{c}
    \includegraphics[width=0.48\textwidth]{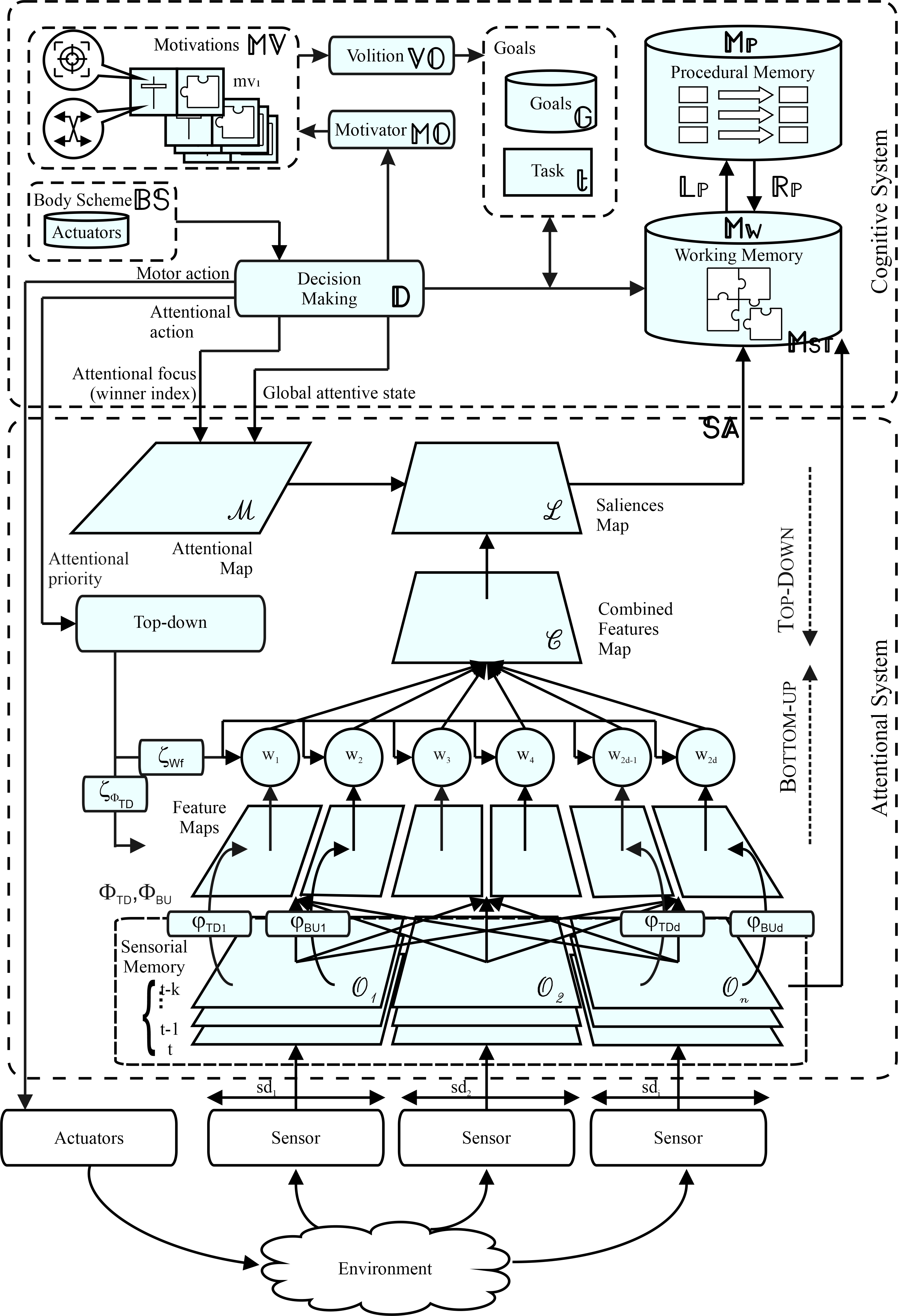} \\ (a) \\
    \includegraphics[width=0.48\textwidth]{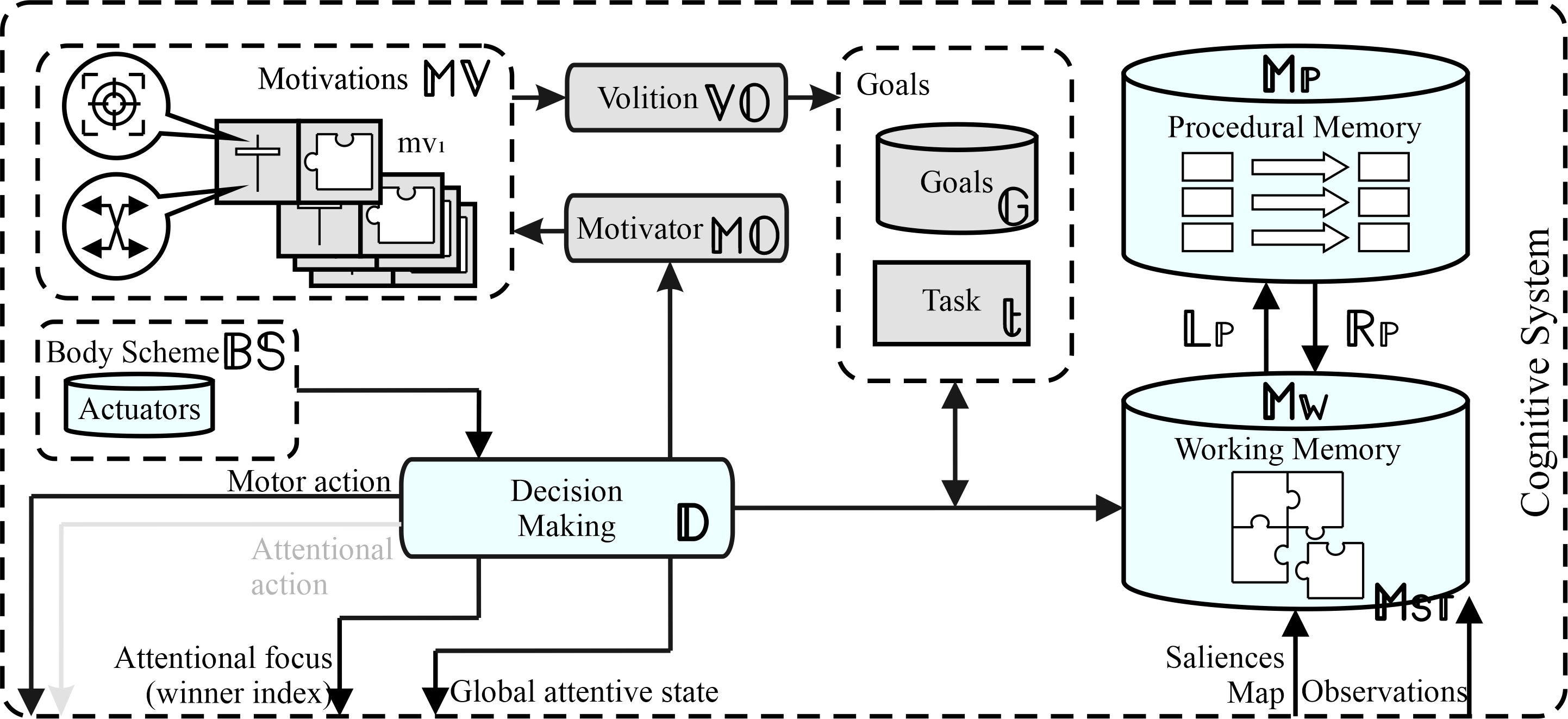}  \\ (b) \\ 
    \includegraphics[width=0.48\textwidth]{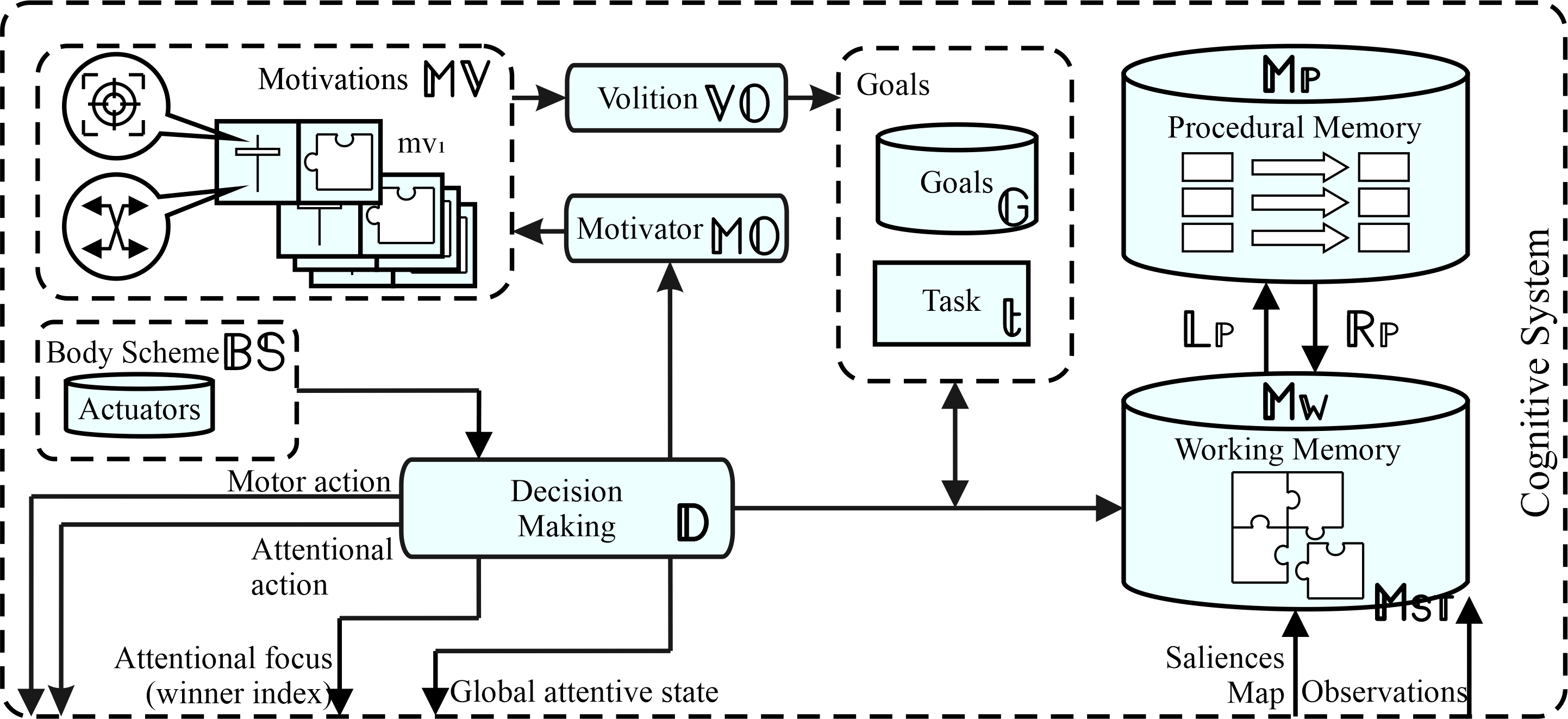} \\ (c) \\
    \end{tabular}
    \end{adjustbox}
    
    \caption{Schematic model of the cognitive-attentive system adopted. a) Full view; b) Details of the Cognitive System in 1st Substage, with some of the modules ($\mathbb{MO}, \mathbb{VO}, \mathbb{MV}, \mathbb{G}, \textbf{t}$, \emph{top-down}) disabled (painted in grey); c) Details of the Cognitive System in 2nd Substage and 3rd Substage.}
    \label{fig:model}
\end{figure*}

\noindent The cognitive system of the humanoid Marta was modeled according to CONAIM \cite{simoes_2015}. A schematic of the adopted cognitive model is shown in Figure \ref{fig:model}. The first level of architecture is the \textbf{attentional system} \cite{colombini2014}, responsible for collecting data from the environment, compressing this information and selecting the most relevant points in the scene. The system's inputs receive multiple sensory information from the four camera channels (R, G, B, D) with a previously configured resolution. These observations generate four distinct \emph{bottom-up} feature maps ($\mathcal{F}_{R}$, $\mathcal{F}_{G}$, $\mathcal{F}_{B}$, $\mathcal{F}_{D}$), which carry information about the most discrepant signals in each channel. Three \emph{top-down} feature maps ($\mathcal{F}_{color}$, $\mathcal{F}_{dist}$, $\mathcal{F}_{reg}$) were also adopted, which can emphasize particular aspects of color, distance and region of the input data according to agent's goals (if applicable). All feature maps are weighted and combined into a single \textit{Combined Feature Map} ($\mathcal{C}$) that carries the information of the most relevant information considering all input data. The \textit{Attentional Map} ($\mathcal{M}$), which carries the information about the attentional focus at time $t-1$ modulated by inhibition of return (IOR) effects over previously selected points, is combined to $\mathcal{C}$ and produces the \textit{Salience Map} ($\mathcal{L}$), which contains the most relevant points of the visual field at present $t$.

As a result, all exogenous stimuli will compete against each other to be the \emph{winner} of this \textit{bottom-up} competitive process, i.e., the most relevant point of the scene. Besides the \emph{bottom-up} process, any particular feature can also be emphasized via an endogenous \emph{top-down} process. Via the top-down pathway, a specific scene region or a desired color can receive the attentional focus depending on the agent's goal.  

\begin{figure*}[ht!]
\begin{adjustbox}{max width=\textwidth}
\centering
  \includegraphics[width=0.9\textwidth]{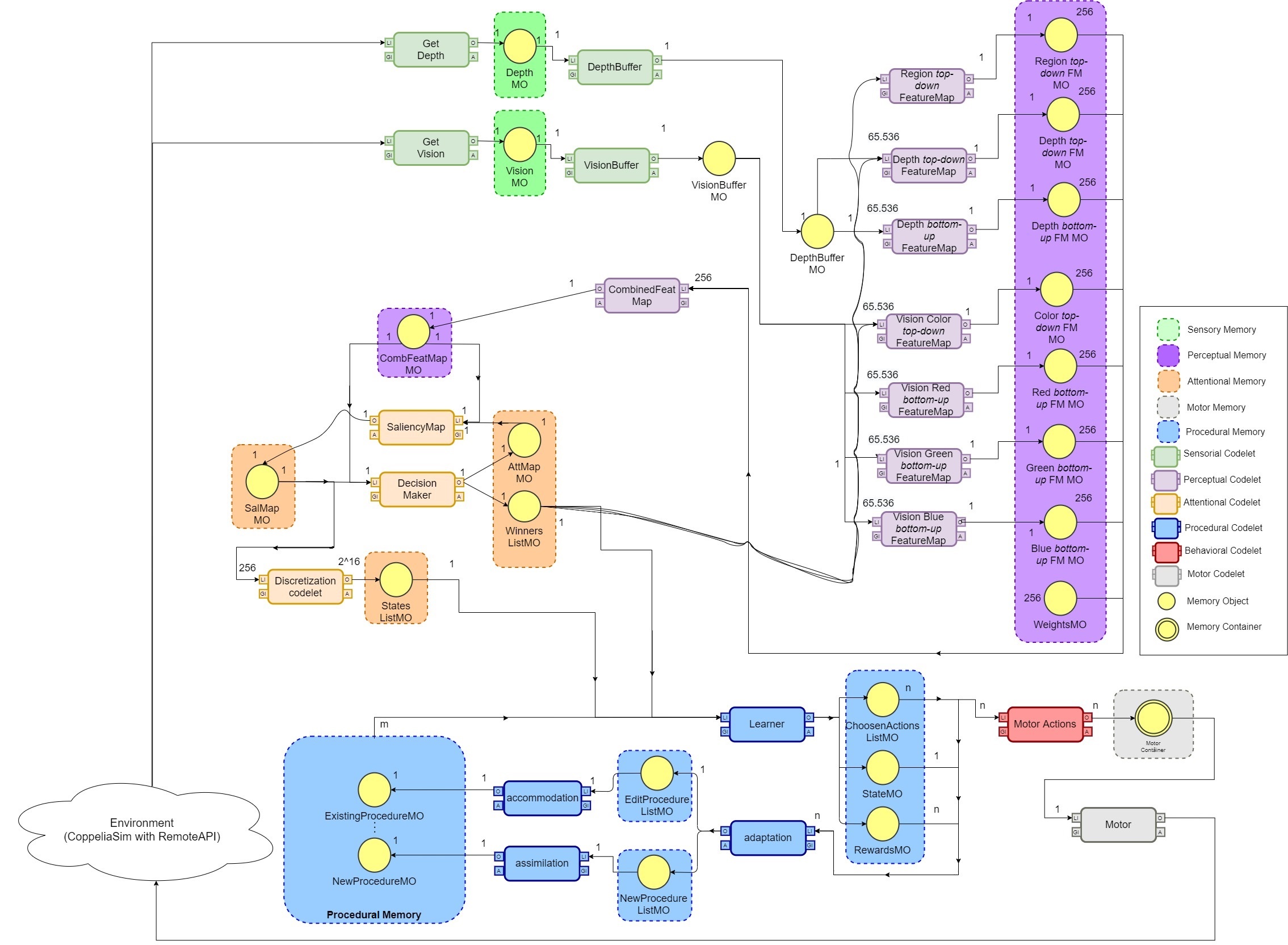}
\end{adjustbox}
\caption{Implementation scheme of the CONAIM+CST architecture with the robot Marta that receives attentional stimuli \emph{bottom-up} and \emph{top-down}.}
\label{fig:scheme}
\end{figure*}

The second level of the architecture is the \textbf{cognitive system}, responsible for modulating the relationship between the robot and the environment, as well as for the cognitive evolution of the agent. In the present experiments, we considered only some modules of CONAIM \cite{simoes_2015} cognitive system. These modules were activated incrementally during successive experiments. Each procedure/schema $m_p \in M_P$ represents learned knowledge stored in procedural memory $\mathbb{M_P}$. Initially, the working memory ($\mathbb{M_W}$) receives the salience map ($\mathcal{L}$) -- used as a state in reinforcement learning -- emerging from the attentional system. As the agent does not know the beginning, a new procedure $m_p$ is created in $\mathbb{M_P}$, and the cognitive agent can gradually learn something about it. If the agent has some prior knowledge stored in $\mathbb{M_P}$ that fits the current state, an \emph{recall} of procedures ($\mathbb{R_P}$) takes place. Decision Maker ($\mathbb{D}$) will consider this knowledge. In some experiments, a set of motivations $mv_i$ $\in \mathbb{M_V}$ was also modeled to explore the use of new actions in some states. The volition $\mathbb{VO}$ is the function responsible for transforming the agent's motivations into tasks that the decision process will also consider.
A procedural learning function ($\mathbb{L_P}$) is responsible for creating or updating the content of $m_p \in M_P$, in this case acting respectively in an analogous way to assimilation and accommodation in Piaget's theory. The cognitive model was fully implemented in \emph{Java} using CST \cite{paraense2016_2}. Figure \ref{fig:scheme} shows the implementation scheme of the CONAIM+CST architecture for the proposed incremental learning.

% Para inserir essa imagem é necessário traduzi-la
%
%\begin{figure*}[ht!]
%\begin{adjustbox}{max width=\textwidth}
%\centering
% \includegraphics[width=0.9\textwidth]{img/model.jpg}
%\end{adjustbox}
%\caption{Humanoid Marta's internal model. The model starts in the 1st sensorimotor substage with only the elements in \textcolor{blue}{blue}. In the 2nd sensorimotor substage, elements in \textcolor{red}{red} are added. In the 3rd sensorimotor substage, elements in \textcolor{orange}{yellow} are added. Source: }
%\label{fig:model}
%\end{figure*}

\subsection{Reinforcement Learning (RL) in substages}
\label{sec:states}

Due to the trial and error nature of learning at early child developmental stages \cite{NUSSENBAUM2019100733}, we adopted RL as the primary paradigm to learn state-action pairs, that is, the agent's procedures in the procedural memory. This section details (1) states, (2) actions, and (3) learning in this approach.

\subsubsection{States} 
Before we can compute the agent State ($\mathbb{S}$), the input of our RL algorithm, we must go back to the sensors data observation ($\mathcal{O}$). Our approach gradually increases the robot's visual acuity, refining the agent's perception of the world. Three different image resolutions were adopted among the experiments: 64x64 pixels (1st Substage), 128x128 pixels (2nd Substage), and 256x256 pixels (3rd Substage). The \emph{bottom-up} maps of the RGB-D channels ($\mathcal{F_R}$, $\mathcal{F_G}$, $\mathcal{F_B}$ and $\mathcal{F_D}$) were computed using an \emph{average pool} over the observation of each color map at time $t$, and then the difference between each region mean and the image mean. Since the resolution of the image changes among the experiments, we computed the necessary size of the kernel and stride to reduce the feature map to a final size of 16x16. In other words, the most discrepant elements at each map at each time $t$ are highlighted. \emph{Top-down Feature Maps} ($\mathcal{F}_{color}$, $\mathcal{F}_{dist}$ and $\mathcal{F}_{reg}$) allow the agent to target its attention to desired elements deliberately. We compare each pixel value to a particular (color, distance, or spatial region) goal to build these maps. The closer these elements are to the target values according to predefined percentage ranges, the higher the map activation in that region. Our experiments adopted 20\%, 40\%, 60\%, and 80\% proximity ranges, respectively. The corresponding attentional values are $1$, $0.75$, $0.5$, and $0.25$. Particularly, in the \emph{regions top-down Feature Map} ($\mathcal{F}_{reg}$), the visual space was divided into $5$ distinct regions, as shown in Figure \ref{fig:Marta} (d). These regions define particular regions of interest for the agent. The \emph{Combined Feature Map} ($\mathcal{C}$) computes an element-wise weighted mean of the $i$ enabled \emph{Features Maps}. An element-wise multiplication of the \emph{Attentional Map} ($\mathcal{M}$) and ($\mathcal{C}$) results in the \emph{Salience Map} ($\mathcal{L}$). Finally, we compute the State ($\mathbb{S}$) vector that will be used as input for the learning algorithm. $\mathbb{S}$ is computed in the \emph{Working Memory} ($\mathbb{M_W}$) using a \emph{MaxPool} operator with a 4x4 kernel and a 4 stride over the Salience Map ($\mathcal{L}$), generating a 4x4 matrix with a 2-level discretization per element obtained with a threshold. This process results in a state vector of size 65.536 ($2^{16}$).

\subsubsection{Actions} 
\begin{figure*}[ht!]
\begin{center}
    \begin{adjustbox}{max width=\textwidth}
        \centering
        \includegraphics[width=0.7\linewidth]{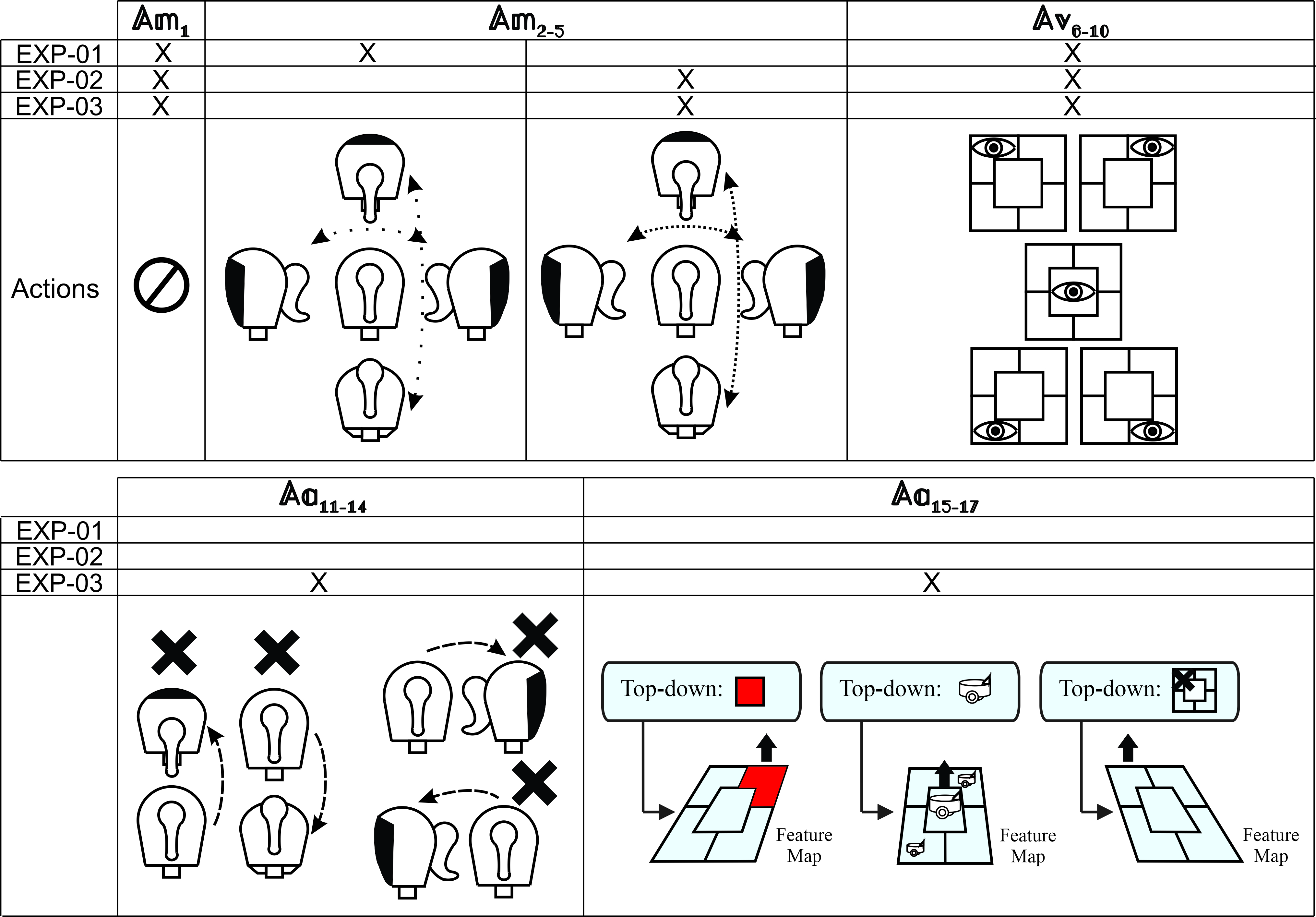}
    \end{adjustbox}
    \caption{Possible actions for the cognitive robot in experiments for 1st Substage, 2nd Substage  and 3rd Substage. \textbf{Motor actions ($\mathbb{A}_m$)}: 1. No-action; 2-5. Move neck pitch/yaw actuators with low discretization or Move neck pitch/yaw actuators with high discretization; \textbf{Virtual actions ($\mathbb{A}_v$)}: 8-10. Move virtual actuators (eyes) to particular image zones; \textbf{Attentional actions ($\mathbb{A}_a$):} 11-14. Move neck pitch/yaw actuators towards the attentional stimulus; \emph{Top-down Attentional Actions}: 15. Emphasize stimuli of a particular color; 16. Emphasize stimuli at a particular distance; 17. Emphasize stimuli in a particular region of the space. }
    \label{fig:actions}
\end{center}
\end{figure*}

In analogy to the typical actions performed by children in each substage, the robot was allowed to perform 17 possible actions ($\mathbb{A}$), divided into three groups: motor, virtual and attentional actions. The Motor Actions ($\mathbb{A}_m$) are the actions on the physical actuators on the robot's neck, capable of turning the head motors \textit{pitch} and \textit{yaw}. Virtual Actions ($\mathbb{A}_v$) are internal to the agent and simulate eye movement. The virtual actuator selects a point in the visual space where the agent focuses (eye). Attention Actions ($\mathbb{A}_a$) are divided into two subgroups. The first group of actions involves directing the robot's head toward the most salient point in the image (winner). The second group refers to \emph{top-down} actions. It can emphasize specific colors, distances, or regions in resource maps. Some of these possible actions have been enabled or disabled for each of the three sets of experiments. Figure \ref{fig:actions} presents the actions available to each distinct substage and experiment. 

\subsubsection{Learning} 

\begin{figure*}[ht!]
    \begin{center}
    \begin{adjustbox}{max width=0.6\textwidth}
        \includegraphics[width=1\linewidth]{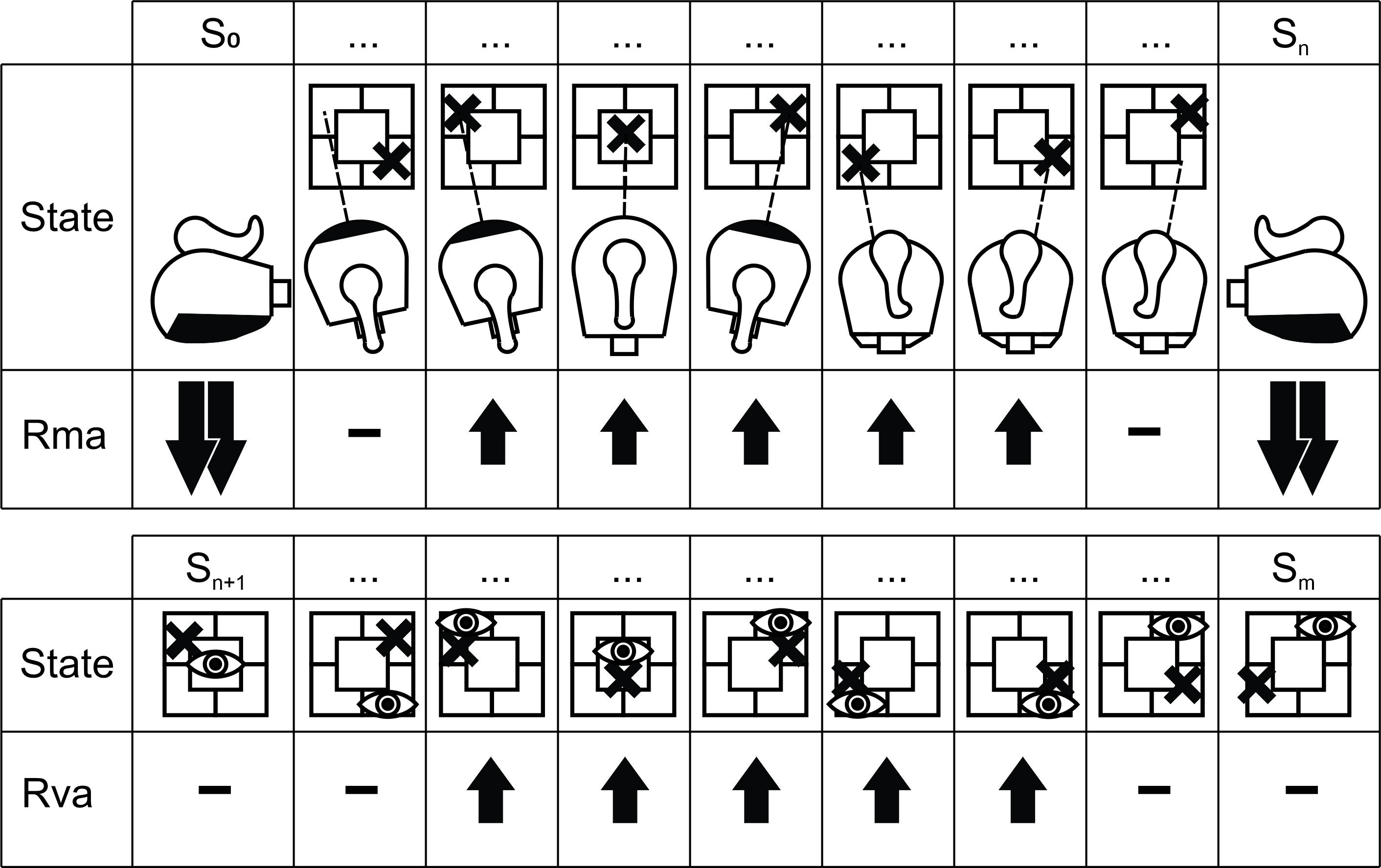}
    \end{adjustbox}
    \caption{Reinforcement for the cognitive robot depends on the current state ($s_i$) and previous action. \textbf{Reinforcement for motor actions (Rma):} The system is positively reinforced if the direction the robot's head moved matches the emergence of a visual stimulus; The system receives no reinforcement if there is no such space-time synchronicity; The system is strongly negatively reinforced if the robot loses its balance; \textbf{Reinforcement for virtual actions (Rva):} The system is positively reinforced if the direction in which the virtual actuator (eye) moved matches the emergence of a visual stimulus; The system receives no reinforcement if there is no such spatial-temporal synchronicity. }
    \label{fig:reinforcement}
    \end{center}
\end{figure*}

The \emph{Learning Proccess} ($\mathbb{L_P}$) has a central role in the current investigation. We selected a \emph{Reinforcement Learning} (RL) algorithm, the Q-learning \cite{sutton1998}, for the cognitive agent's learning. The memory elements $m_p \in \mathbb{M_P}$ were modeled as \emph{QTables} capable of storing State-Action pairs ($\mathbb{S} \rightarrow \mathbb{A}$) for particular procedures. The states ($\mathbb{S}$) were modeled from the saliency maps ($\mathcal{L}$) that represent the environment. Reinforcement positively rewards the robot if there is space-time synchronization between the visual stimulus (most salient point of the image) and the robot's current focus (motor or virtual). There is no reward if there is no such timing, and the reward is strongly negative if the robot loses balance. The learning mechanism remains unchanged during all experiments. Figure \ref{fig:reinforcement} details the reinforcement policy for some states of the agent.

\subsection{Proposed Experiments}

\noindent Three sets of experiments (EXP-01/03) were proposed based on the scenarios to assess the agent's ability to learn to track other agents or objects, proposed in section \ref{sec:proposed_approach} \cite{berto_2020_thesis}. During training, for each episode, we initialize the agent in a fixed position in the environment, adding random noise to its actuators. 

The episode ends when one of the following conditions is reached: \textit{i)} the agent reached the maximum number of steps/actions; \textit{ii)} the robot falls over or exceeds the limits of its motorized actuators, or \textit{iii)} the robot has no saliences for several iterations. The following parameters were adopted:

\begin{itemize}
    \item \textbf{Simulation parameters.} Maximum number of episodes: 200; Maximum number of steps: 500; Maximum number of iterations without bosses: 5;
    \item \textbf{Learning parameters.} A $\epsilon-$greedy policy was employed with an exploitation rate starting at 0.95 and linearly decaying to 0 in the last episode. Learning rate $\alpha$: 0.9; Time discount rate $\gamma$: 0.99;
    \item \textbf{Visual acuity / Vision sensor resolutions.} 64x64 for $1^{st}$ substage; 128x128 for $2^{nd}$ substage; 256x256 for $3^{rd}$ substage;
    \item \textbf{Rewards.} +1 reward for each new data inserted in procedural memory; +1 for holding or directing an actuator (motor or virtual) to the attention winner; -10 if the agent falls or exceeds the limits of physical actuators, or if the agent has no saliences in its attentional cycle for several iterations; and, only for $3^{rd}$ substage, +1 when the agent identifies regions in which a certain desired characteristic is highlighted according to the \emph{top-down} process.
\end{itemize}

Table \ref{tab:detailed} details the experiments carried out, the cognitive system, available modules, functions and learning.

\begin{table*}[ht!]
    \begin{center}
    \begin{scriptsize}
    \begin{tabular}{ | m{0.4cm} | m{0.4cm} |  m{0.7cm} | m{0.8cm}| m{2.2cm} | m{2.5cm} | m{3.5cm} | m{4.1cm} |} 
        \hline
        EXP & Substage & Vision resolution & Bottom-up features & Memories & Actions & Cognitive functions & RL Reward strategy \\ 
        \hline
        01 & 1 & 64x64 pixels & RGBD & $\bullet$ Short-term: Sensory Memory ($\mathbb{M_S}$) and Working Memory ($\mathbb{M_W}$) \newline $\bullet$ Long-term: Procedural Memory ($\mathbb{M_P}$) & $\bullet$ Motor actions ($\mathbb{A}_m$): stay focused; move physical actuators \newline $\bullet$ Virtual action ($\mathbb{A}_v$): move virtual actuator \newline $\bullet$ Attentional actions ($\mathbb{A}_a$): - & $\bullet$ Procedural Learning ($\mathbb{L_P}$): Q-Learning \newline $\bullet$ Motivation ($\mathbb{MO}$): - & The agent receives a +1 reward for each new data inserted into the procedural memory and also when it directs its face (physical) or eyes (virtual) to the winner of the attentional process. The agent receives a -10 reward if it falls, exceeds the actuator limits, or does not have saliences in its attentional cycle for a number of iterations \\
        \hline
        02 & 2 & 128x128 pixels & RGBD & $\bullet$ Short-term: Sensory Memory ($\mathbb{M_S}$) and Working Memory ($\mathbb{M_W}$) \newline $\bullet$ Long-term: Procedural Memory ($\mathbb{M_P}$) & $\bullet$ Motor actions ($\mathbb{A}_m$): stay focused; move physical actuators \newline $\bullet$ Virtual action ($\mathbb{A}_v$): move virtual actuator \newline $\bullet$ Attentional actions ($\mathbb{A}_a$): - & $\bullet$ Procedural Learning ($\mathbb{L_P}$): Q-Learning \newline $\bullet$ Motivation ($\mathbb{MO}$): Motivation modifies the curiosity drive's activation. This driver incentives actions not tried in a particular state. Hence, encouraging new sensorimotor discoveries & same as previous\\
        \hline
        03 & 3 & 256x256 pixels & RGBD & $\bullet$ Short-term: Sensory Memory ($\mathbb{M_S}$) and Working Memory ($\mathbb{M_W}$) \newline $\bullet$ Long-term: Procedural Memory ($\mathbb{M_P}$) & $\bullet$ Motor actions ($\mathbb{A}_m$): stay focused; move physical actuators \newline $\bullet$ Virtual action ($\mathbb{A}_v$): move virtual actuator \newline $\bullet$ Attentional actions ($\mathbb{A}_a$): direct attentional focus according to features \textit{top-down} & $\bullet$ Procedural Learning ($\mathbb{L_P}$): Q-Learning \newline $\bullet$ Motivation ($\mathbb{MO}$): Motivation modifies the curiosity drive's activation. This driver incentives actions not tried in a particular state. Hence, encouraging new sensorimotor discoveries & The agent receives +1 reward for each new data inserted in the procedural memory; +1 for keeping or directing an actuator (motor or virtual) to the winner of the attention; +1 when the agent identifies regions in which a certain desired feature is highlighted according to the top-down process; -10 for falling, motor actuators limits exceed, or if it does not have saliences in its attentional cycle for several iterations \\
        \hline
    \end{tabular}
    \end{scriptsize}
    \end{center}
    \caption{Details of the experiments: substages, vision resolution, memories, actions, cognitive functions and reward strategy.}
    \label{tab:detailed}
\end{table*}

\subsubsection{1st Substage: Use of Reflexes} 

\begin{figure*}[ht!]
    \centering
    \begin{adjustbox}{max width=\textwidth}
    \includegraphics[width=1\linewidth]{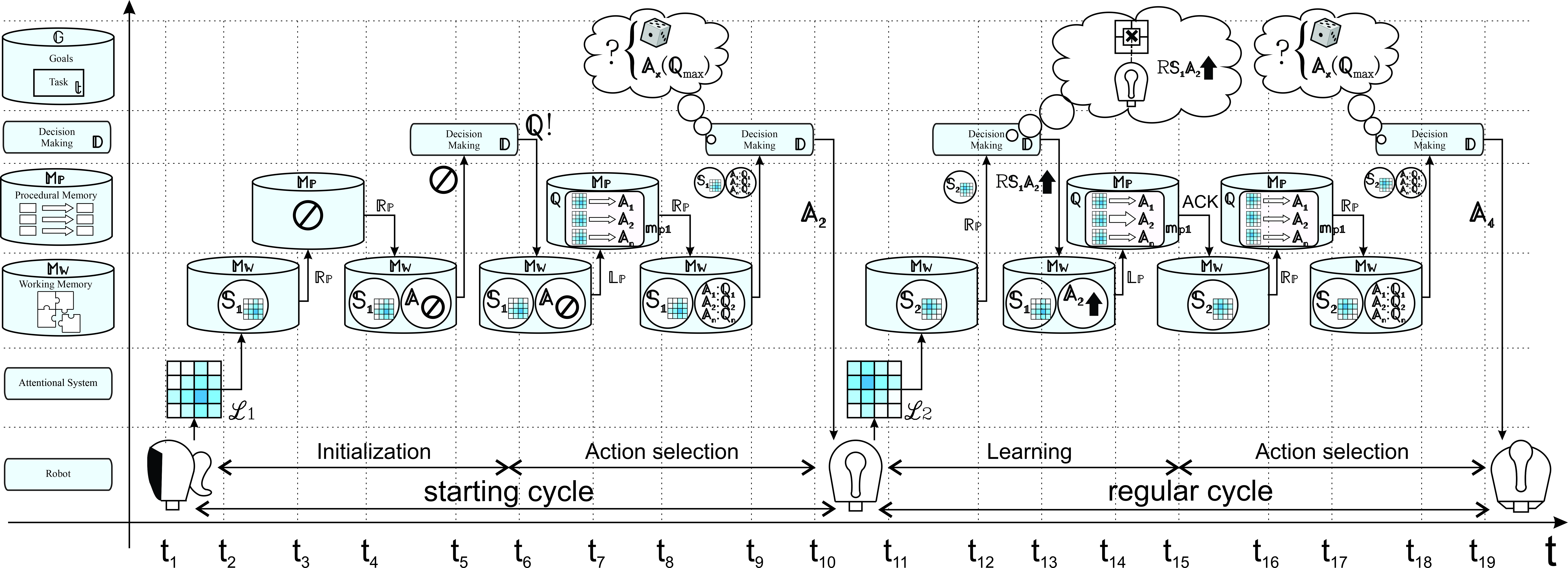}
    \end{adjustbox}
    \caption{System dynamics in 1st Substage. $t_1$: The robot sensors sample the environment; $t_2$: The attentional system generates the saliency map ($\mathcal{L}_1$); $t_3$: The working memory ($\mathbb{M_W}$) identifies this new state of the world ($\mathbb{S}_1$) based on ($\mathcal{L}_1$). The recall function ($\mathbb{R_P}$) is called to look for for any procedure in $\mathbb{M_P}$ that could be applicable to the current state $\mathbb{S}_1$; $t_4$: Such procedure is not found in $\mathbb{M_P}$; $t_5$: The decision maker ($\mathbb{D}$) informs that the robot do not know what to do in the current state; $t_6$: The decision maker ($\mathbb{D}$) decides to start a new QTable associated to this state ($\mathbb{S}_1$) in $\mathbb{M_P}$; $t_7$: The new QTable is created with random values, enabling the cognitive robot to learn more about this state; $t_8$: The recall function $\mathbb{R_P}$ returns the possible actions associated to this state and their $\mathbb{Q}$ values; $t_9$: current state $\mathbb{S}_1$ and possible actions are sent to the decision maker ($\mathcal{D}$); $t_{10}$: The decision maker ($\mathcal{D}$) chooses the next action between a random action and the best action for the current state ($\mathcal{Q}_{max}$) according to an $\epsilon$-\emph{greedy} policy. In this example, the action to move the robot's head to the right ($\mathbb{A}_2$) was selected; $t_{11}$: A new salience map ($\mathcal{L}_2$) is generated by the attentional system; $t_{12}$: The working memory ($\mathbb{M_W}$) identifies the new state ($\mathbb{S}_2$) based on ($\mathcal{L}_2$); $t_{13}$: The decision maker ($\mathcal{D}$) evaluates the current state ($\mathbb{S}_2$) and the previous action performed ($\mathbb{A}_2$). Since there is a synchronicity between the robot's head-resulting position and the external stimulus, a positive reinforcement $R_{\mathbb{S}_1\mathbb{A}_2}$ increases the $\mathbb{Q}$ value for the action $\mathbb{A}_2$ in the state $\mathbb{S}_1$; $t_{14}$: The $\mathbb{Q}$ value is sent to the QTable; $t_{15}$: The QTable is updated; $t_{16}$: A new recall $\mathbb{R_L}$ looks for the possible actions for the current state $\mathbb{S}_2$; $t_{17}$: The possible actions are found in $\mathbb{M_P}$; $t_{18}$: The decision maker ($\mathbb{D}$) is informed about the current state $\mathbb{S}_2$ and the possible actions; $t_{19}$: The decision maker ($\mathbb{D}$) decides by the action $\mathbb{A}_4$ (move the head bellow). }
    \label{fig:exp-01}
\end{figure*}

The 1st Substage experiments dynamics is schematically presented in Figure \ref{fig:exp-01}. This set of experiments investigates a computational process proposed to model \textbf{reflex reactions}. In this set, there is no intentionality or motivation.

%\textbf{Cognitive elements:}
%\begin{itemize}
%    \item \textbf{Sensors:} vision;
%    \item \textbf{Discretization/resolution:} 64 x 64 pixels, in an analogy to the child's low visual accuracy, which will be gradually increased in the next stages;
%    \item \textbf{Features \emph{bottom-up}:} color channel intensities (RGB) and relative distance;
%    \item \textbf{Memories:} \textbf{Short-term:} Sensory Memory and Working Memory. \textbf{Long-term:} Procedural Memory;
%    \item \textbf{Motor actions:} stay focused; move physical actuators; and move the virtual actuator;
%    \item \textbf{Procedural Learning:} implemented through the reinforcement learning algorithm Q-Learning;
%    \item \textbf{Rewards:} The agent receives a +1 reward for each new data inserted into the procedural memory and also when it directs its face (physical) or eyes (virtual) to the winner of the attentional process. The agent receives a -10 reward if it falls, exceeds the actuator limits, or does not have saliences in its attentional cycle for a number of iterations. 
%\end{itemize}

\subsubsection{2nd Substage: Primary Circular Reactions} 

In this set of experiments, we investigated whether the reflex reactions in the agent, initiated in the 1st substage, can evolve into behaviors similar to the primary circular reactions proposed by Piaget. As in the previous experiment, only the \emph{bottom-up} stimuli were considered. \textbf{Motivation.} A motivation function ($\mathbb{MO}$) is applied, related to the agent's curiosity about the effects of its actions, to encourage the agent to explore schemes that have not yet been explored in the current episode. All previously learned contents of $\mathbb{M_P}$ are preserved.

%\textbf{Cognitive elements:}
%\begin{itemize}
%    \item \textbf{Sensors:} vision;
%     \item \textbf{Discretization/resolution:} 128 x 128 pixels;
%    \item \textbf{Features \emph{bottom-up}:} color channel intensities (RGB) and relative distance;
%    \item \textbf{Memories:} \textbf{Short-term:} Sensory Memory and Working Memory. \textbf{Long-term:} Procedural Memory;
%    \item \textbf{Motor actions:} stay focused; move physical actuators; and move the virtual actuator;
%    \item \textbf{Curiosity and motivation:} %the possibility of reusing knowledge from the previous substage allows the agent to develop a curiosity drive, encouraging or suppressing the execution of specific actions, depending on its performance. 
%    The motivation modifies a curiosity drive's activation. This driver incentives actions not tried in a particular state. Hence, encouraging new sensorimotor discoveries;
%    \item \textbf{Procedural Learning:} implemented through the reinforcement learning algorithm Q-Learning;
%    \item \textbf{Rewards:} same as previous substage.
%\end{itemize}

\subsubsection{3rd Substage: Secondary Circular Reactions} In this last set of experiments, we investigate the computational process by which behaviors associated with the secondary circular reactions proposed by Piaget can be observed. We verified whether the agent could intentionally select an action that would allow him to reach a goal. The \emph{top-down} attention mechanism is used in this phase. All previously learned contents of $\mathbb{M_P}$ are preserved.

%\textbf{Cognitive elements:}
%\begin{itemize}
%    \item \textbf{Sensors:} vision;
%    \item \textbf{Discretization/resolution:} 256 x 256 pixels;
%    \item \textbf{Features \emph{bottom-up}:} color channel intensities (RGB) and relative distance;
%    \item \textbf{Features \emph{top-down}:} desired-color, desired-distance and desired-region;
%    \item \textbf{Memories:} \textbf{Short-term:} Sensory Memory and Working Memory. \textbf{Long-term:} Procedural Memory;
%    \item \textbf{Motor actions:} stay focused; move physical actuators; and move the virtual actuator;
%    \item \textbf{Attentional actions:} direct attentional focus according to features \emph{top-down};
%    \item \textbf{Curiosity and motivation:} the agent remains with its curiosity drive, being able to encourage or suppress the execution of certain actions, depending on its performance;
%    \item \textbf{Procedural Learning:} implemented through the reinforcement learning algorithm Q-Learning;
%    \item \textbf{Rewards:} +1 reward for each new data inserted in the procedural memory; +1 for keeping or directing an actuator (motor or virtual) to the winner of the attention; +1 when the agent identifies regions in which a certain desired feature is highlighted according to the \emph{top-down} process; -10 for falling, motor actuators limits exceed, or if it does not have saliences in its attentional cycle for several iterations.
%\end{itemize}

\section{Results and Discussion}

\begin{figure*}[ht!]
     \centering
 \includegraphics[width=0.49\linewidth]{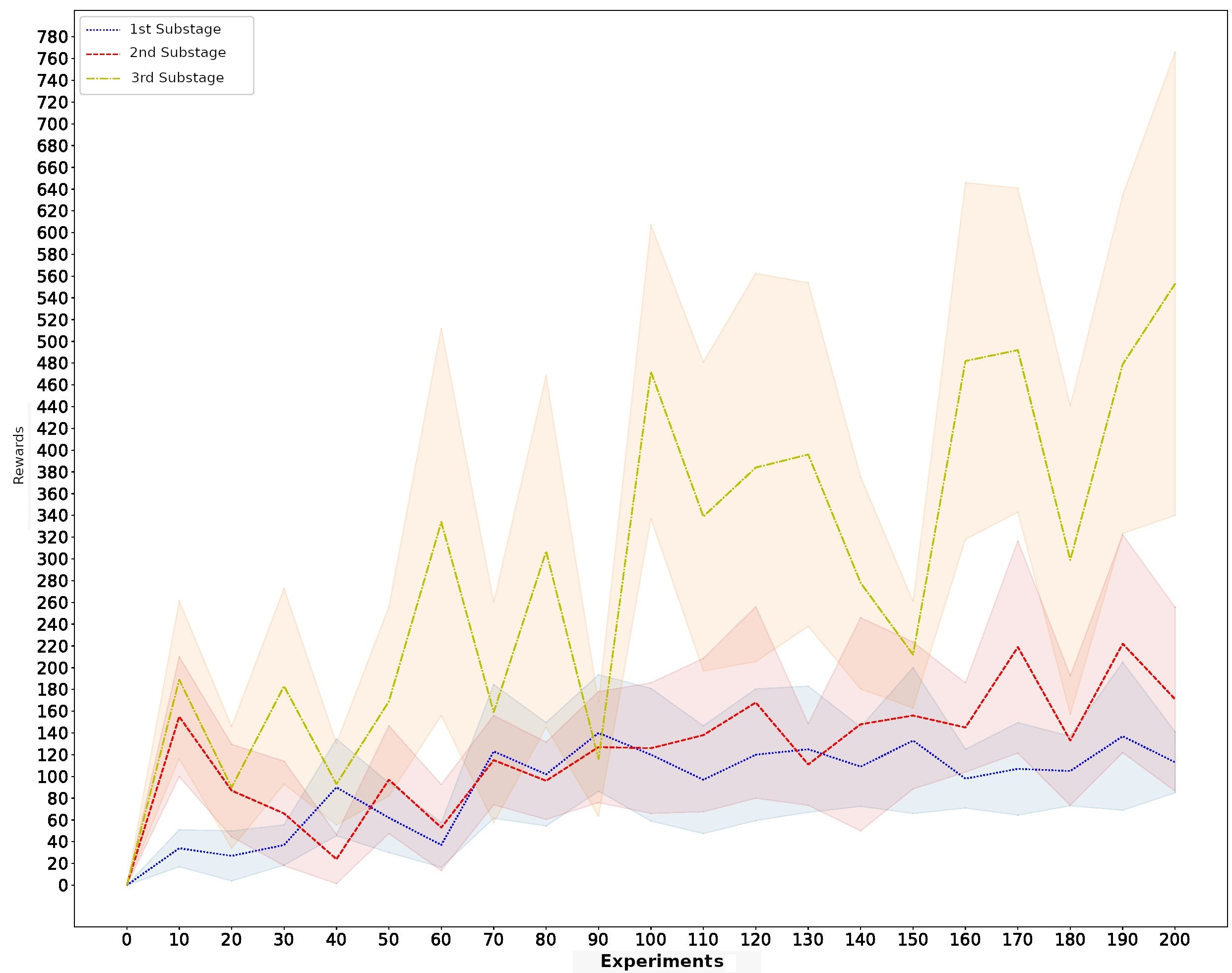}
     \includegraphics[width=0.49\linewidth]{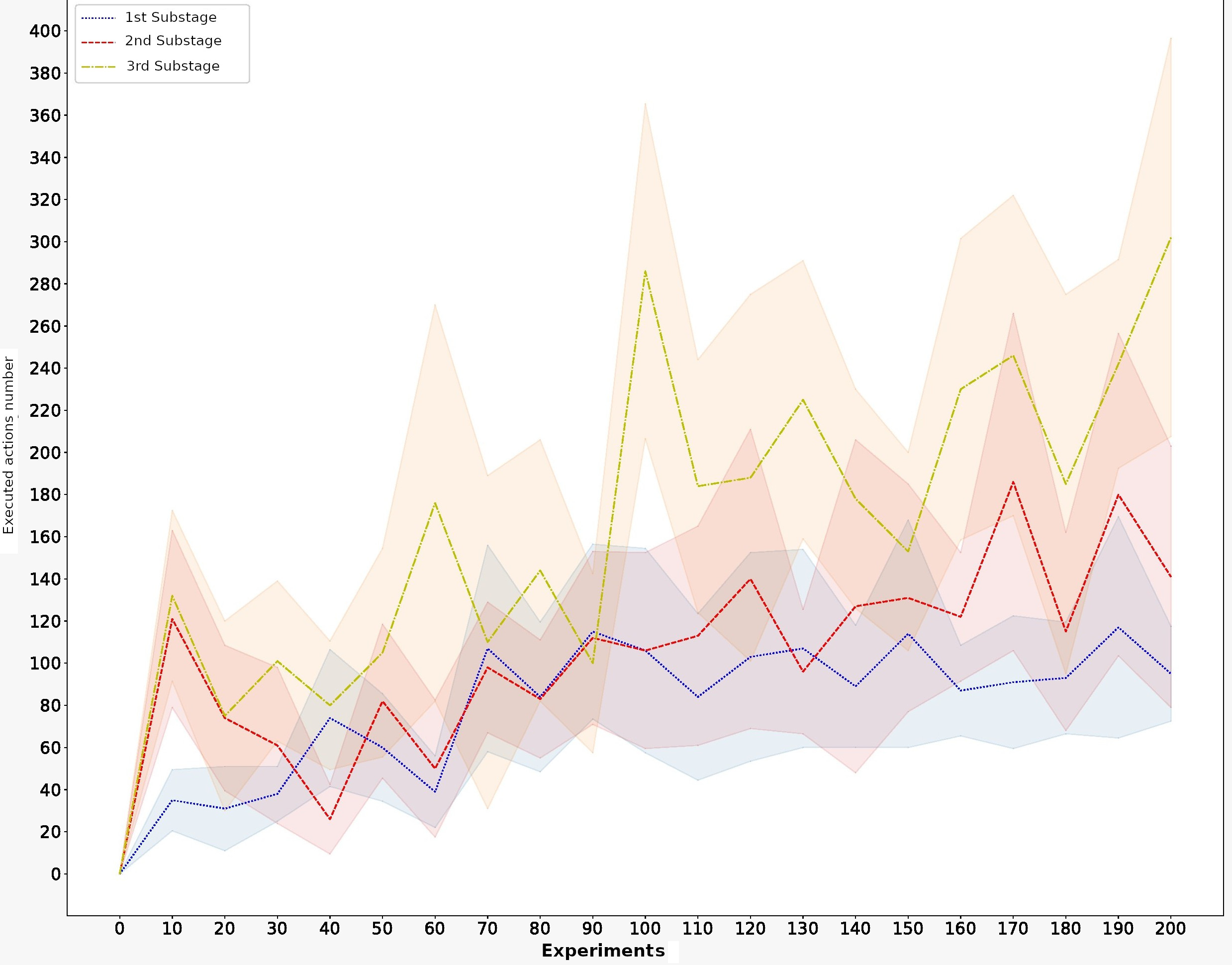}
     \includegraphics[width=0.49\linewidth]{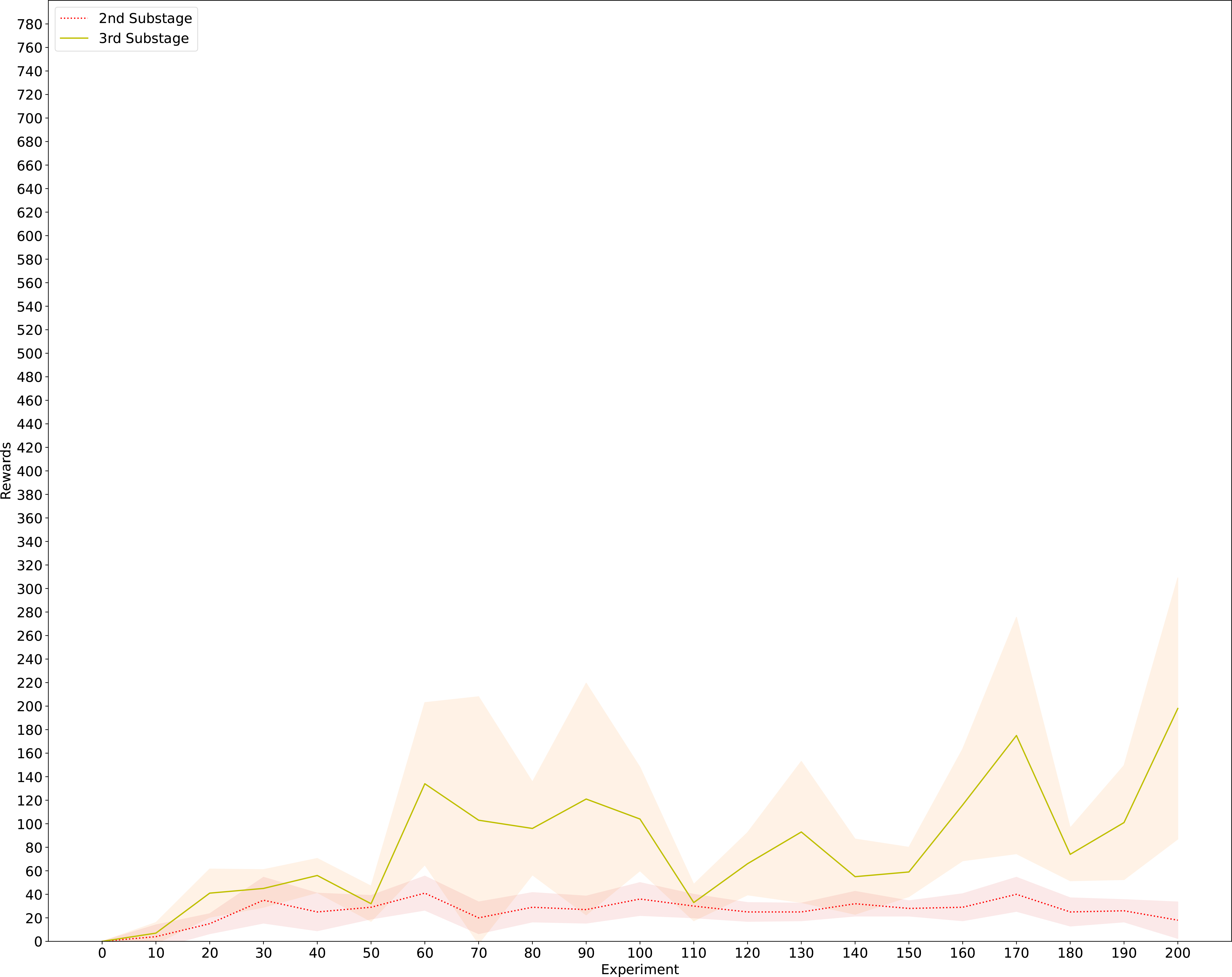}
     \includegraphics[width=0.49\linewidth]{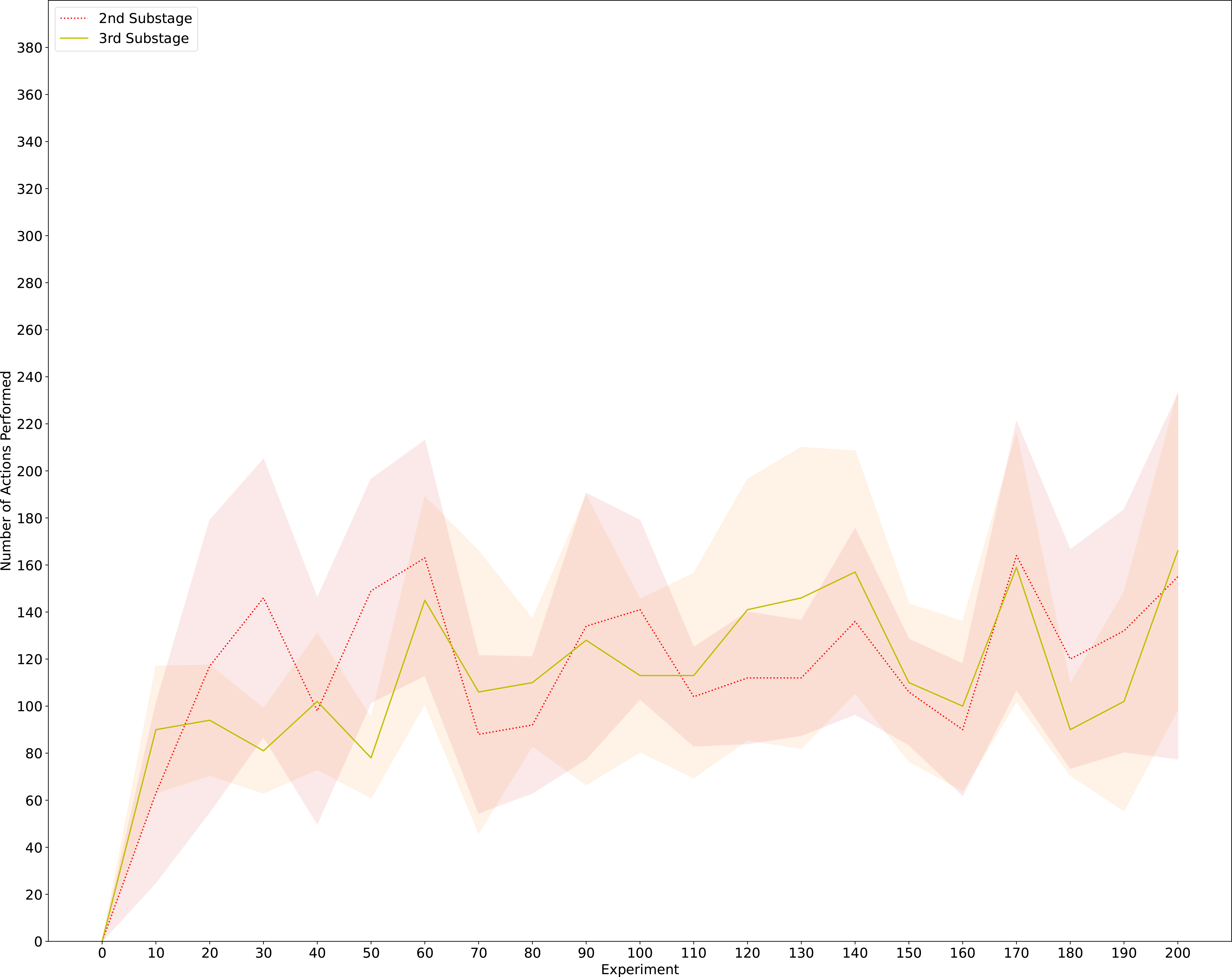}
     
     \caption{Top: Resulting reward (left) per episode and number of actions (right) for each learning experiment for all substages when incremental learning is in course. Bottom: Resulting reward (left) per episode and number of actions (right) for each learning experiment for all substages with no incremental learning.}
     \label{fig:results_actions}

\end{figure*}

\noindent We carried out three experiment sets to train and validate the integrated architecture, corresponding to the first three substages of the sensorimotor period in Piaget's Theory \cite{piaget_origins_1952}. At the end of each training episode, the reward obtained and the number of actions performed were restarted and the robot actuators returned to the starting position. The Pioneer P3DX robot was randomly positioned in the scene. Figure \ref{fig:results_actions} demonstrates the resulting reward and the number of actions performed per episode for each learning experiment performed. It can be noted from the top graphs that, as the agent reuses knowledge from previous substages, both the reward and the number of actions are greater for a more developed agent ($3rd > 2nd > 1st$). The bottom images depict the training results when we do not reuse knowledge from prior stages. We can note that, in these cases, either no learning occurs (the reward for the 2nd substage experiment does not increase over episodes) or it results in very low rewards when compared to the scenario where knowledge was reused (3rd substage - reward peak around 200 versus 500). 

\subsection{1st Substage: Use of Reflexes.} 
\begin{figure*}[!h]
    \centering
    \begin{adjustbox}{max width=\textwidth}
     \begin{tabular}{ccccc}
    \includegraphics[height=2.2 cm]{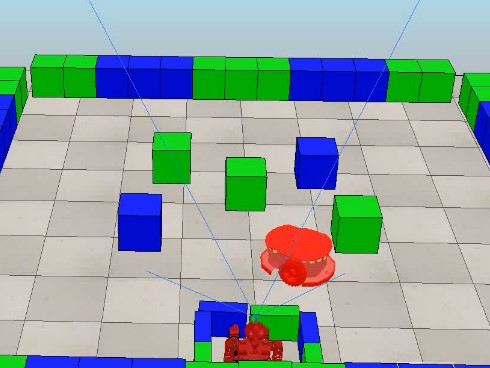}&
    \includegraphics[height=2.2 cm]{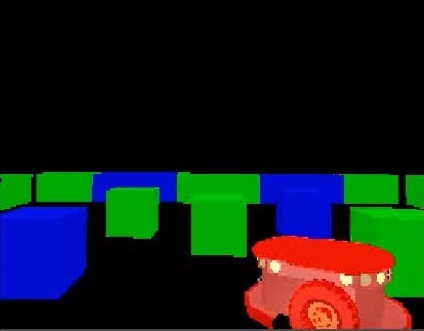}&
     \includegraphics[height=2.3cm]{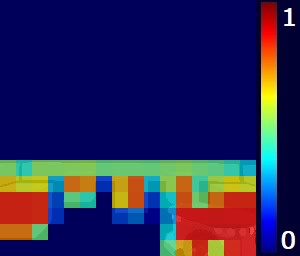}&
    \includegraphics[height=2.3cm]{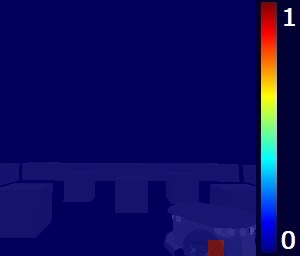}&
    \includegraphics[height=2.5cm]{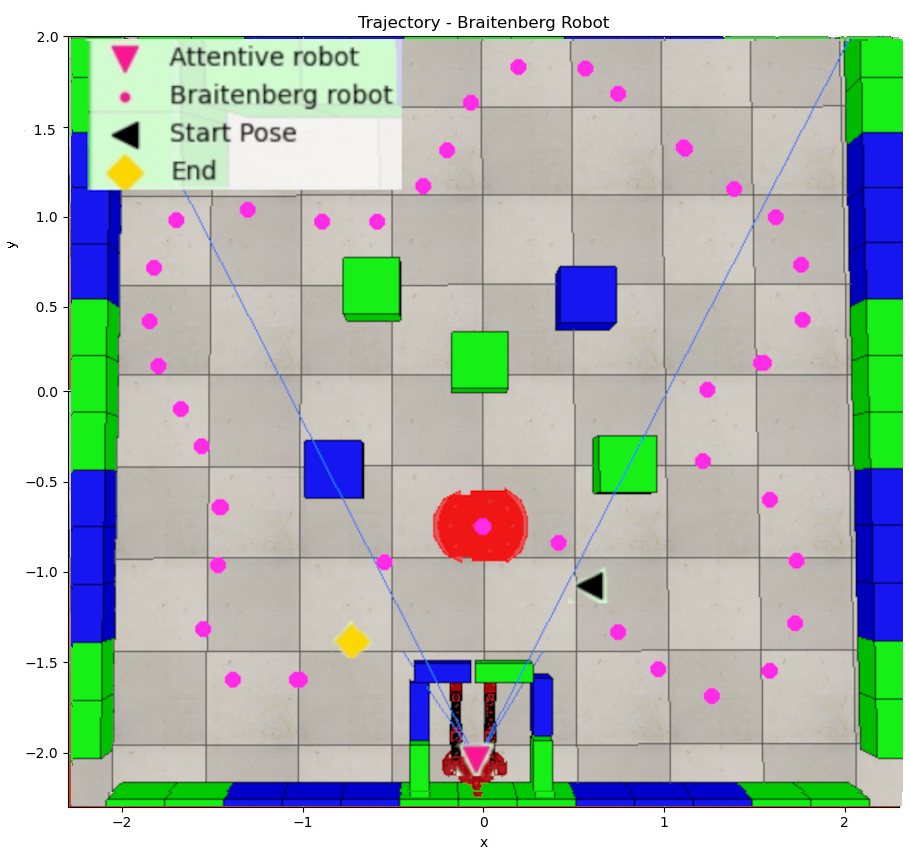}
    \\ \\
    
%    (a) Cena & (b) Sensor de visão & (c) Mapa de saliência &  (d) Vencedor \\

    \end{tabular}
    \end{adjustbox}
    \caption{1st Substage. Sensory data obtained in the 1st episode of Procedural Learning. Left to Right:
     (a) Overview of the scene in \emph{CoppeliaSim} (t = 1s);
     (b) Marta's camera view (t = 1s);
     (c) Salience Map (t = 3s);
     (d) Winner of the attentional cycle  (t = 3s);
     (e) Agents and objects' positions in scene.}
    \label{fig:exps_st1_att}

\end{figure*}

\noindent The experiment demonstrates the attentional course of selection for perception for the agent executing only the \emph{bottom-up} attentional course. The results obtained for Procedural Learning (training) in this experiment are shown in Figure \ref{fig:exps_st1_att}. In the first episodes, the agent explores the limits of its actuators due to the high exploration rate while promoting the refinement of state-action pairs for the fovea selection virtual actuator. The agent established its attentional focus on the Pioneer P3DX robot while it moved in regions closer to the humanoid. With the distance from the Pioneer P3DX robot, the agent directed its attention to other nearby objects and its own body. The stimuli obtained while exploring the agent's body reinforced the reflexes used. The absence of a motivation system made the agent perform the reinforced actions in greater quantity, even when the stimuli that promoted this reinforcement were no longer present and in smaller quantities the actions that did not participate in these interactions.

\noindent \textbf{Learning Validation - 1st Substage:} 
For the validation of the agent in this substage, the QTable resulting from the end of the last episode of Procedural Learning was used. To evaluate the learned policy, 100 test episodes were executed without updating the learning parameters for each experiment, with a maximum of 500 actions.

\subsubsection{\textbf{Experiment A - 1st Substage - Object in fixed position and with primary color}} 
\begin{figure*}[!h]
    \centering
    \begin{adjustbox}{max width=\textwidth}
     \begin{tabular}{ccccc}
    \includegraphics[height=2.2 cm]{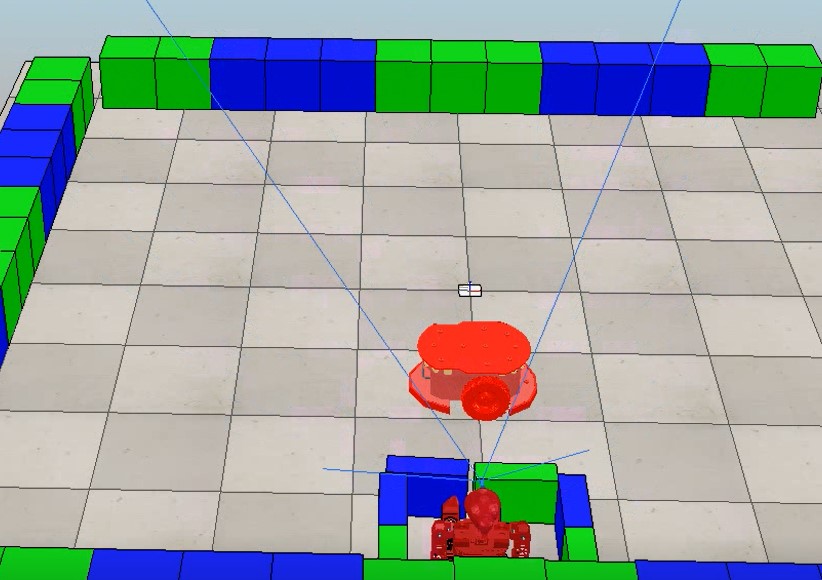}&
    \includegraphics[height=2.2 cm]{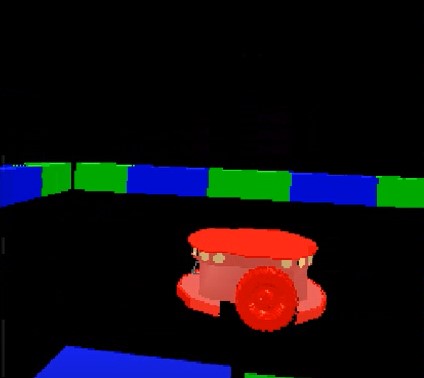}&
     \includegraphics[height=2.3cm]{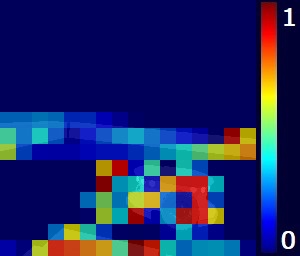}&
    \includegraphics[height=2.3cm]{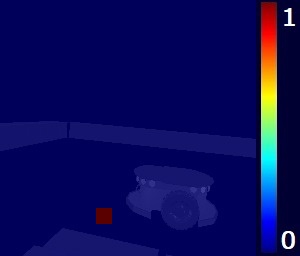} & 
    \includegraphics[height=2.5cm]{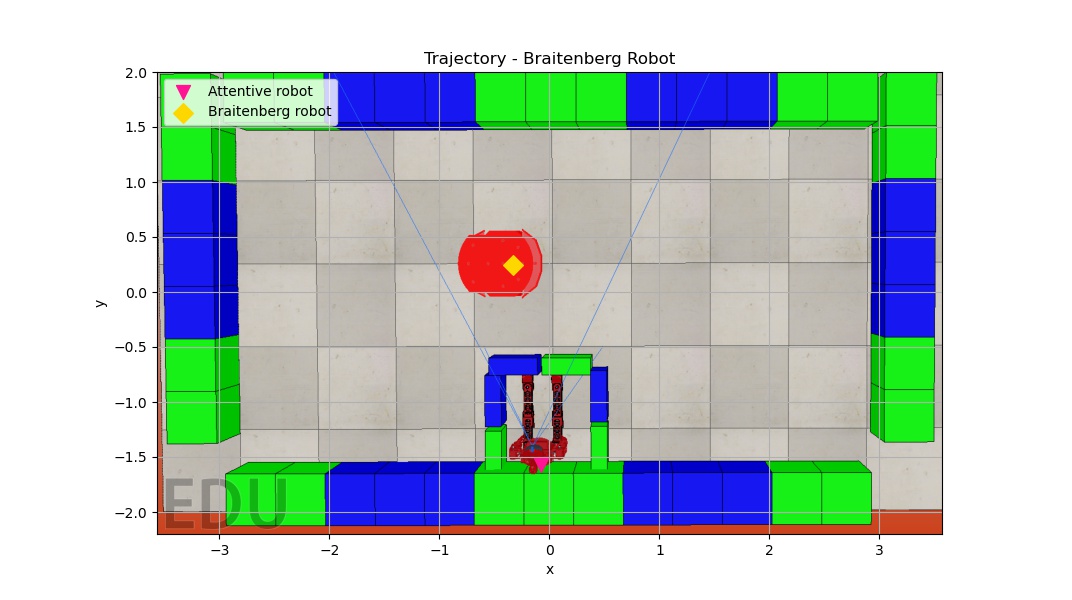} \\ \\
    
   % (a) Cena & (b) Sensor de visão & (c) Mapa de saliência &  (d) Vencedor \\

    \end{tabular}
     \end{adjustbox}
    \caption{1st Substage. Sensory data obtained in validation Experiment A. Left to right:
     (a) Overview of the scene in \emph{CoppeliaSim} (t = 40s);
     (b) Marta's camera view (t = 40s);
     (c) Salience Map (t = 43s);
     (d) Winner of the attentional cycle  (t = 43s);
     (e) Agents and objects' positions in scene.}
    \label{fig:exp_berto_sub1_a}

\end{figure*}

The humanoid Marta was positioned 80cm from the Pioneer P3DX robot. The results obtained for this experiment are shown in Figure \ref{fig:exp_berto_sub1_a}. The agent initially directed its attention to the Pioneer P3DX robot, which remained stationary during this experiment, as suggested by Berto (2020). \cite{berto_2020_thesis}. However, the action of excitatory and inhibitory cycles promoted by the CONAIM attentional system directed the attentional focus to regions closer to the humanoid. The performance of the reinforced reflexes during the exploration of the agent's body in Procedural Learning resulted in the alternation of the agent's actuators between the regions closest to the humanoid and the Pioneer P3DX robot. As expected, the robot learned to respond to salient stimuli using reflexes.

\subsubsection{ \textbf{Experiment B - 1st Substage - Object moving slowly and of primary color}} 
\begin{figure*}[!h]
    \centering
    \begin{adjustbox}{max width=\textwidth}
     \begin{tabular}{ccccc}
    \includegraphics[height=2.2 cm]{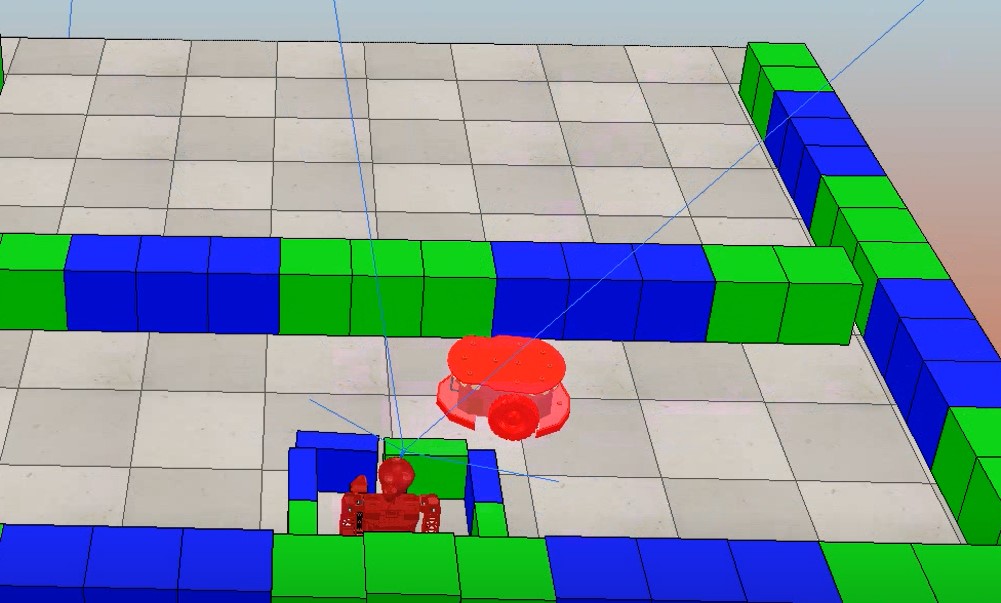}&
    \includegraphics[height=2.2 cm]{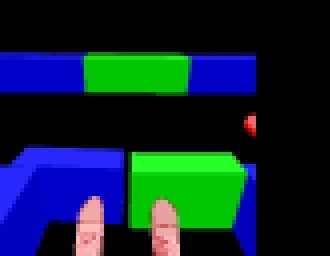}&
     \includegraphics[height=2.3cm]{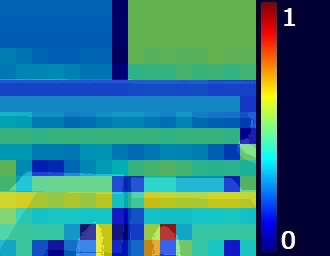}&
    \includegraphics[height=2.3cm]{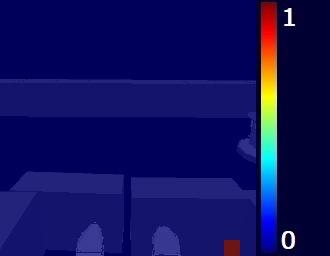}& 
    \includegraphics[height=2.5cm]{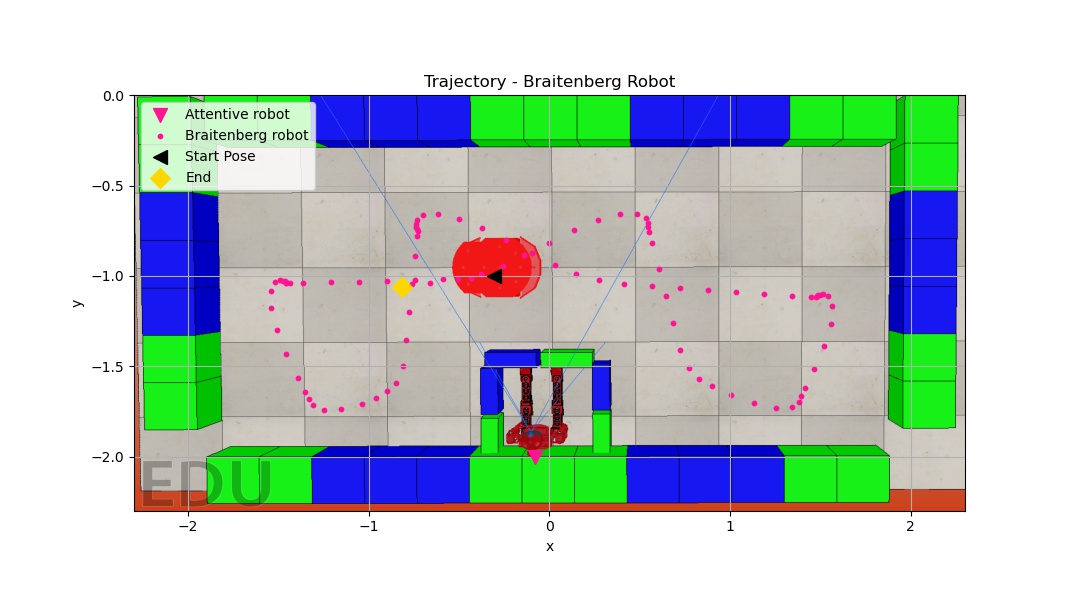}\\ \\
%    (a) Cena & (b) Sensor de visão & (c) Mapa de saliência &  (d) Vencedor \\

    \end{tabular}
     \end{adjustbox}
    \caption{1st Substage. Sensory data obtained in validation Experiment B. Left to right:
     (a) Overview of the scene in the simulator \emph{CoppeliaSim} (t = 40s);
     (b) Marta's camera view (t = 40s);
     (c) Salience Map (t = 43s);
     (d) Winner of the attentional cycle (t = 43s);
     (e) Agents and objects' positions in scene.}
    \label{fig:exp_berto_sub1_b}

\end{figure*}

Marta was positioned 80cm in front of the Pioneer P3DX robot, which has a Braitenberg algorithm and moves at a constant speed of 0.1m/s. The results of this experiment are shown in Figure \ref{fig:exp_berto_sub1_b}. The agent directed its attention to the Pioneer P3DX robot while it remained in the regions closest to the agent (frontal region). However, as the Pioneer P3DX moves to the lateral areas, the performance of the reinforced reflexes during Procedural Learning again resulted in directing attentional focus to the regions closer to the humanoid. Thus, Marta could not track the moving object outside its visual field, as expected for this developmental stage.

\subsection{2nd Substage: Primary Circular Reactions.} 
\begin{figure*}[!h]
    \centering
    \begin{adjustbox}{max width=\textwidth}
     \begin{tabular}{ccccc}
    \includegraphics[height=2.2 cm]{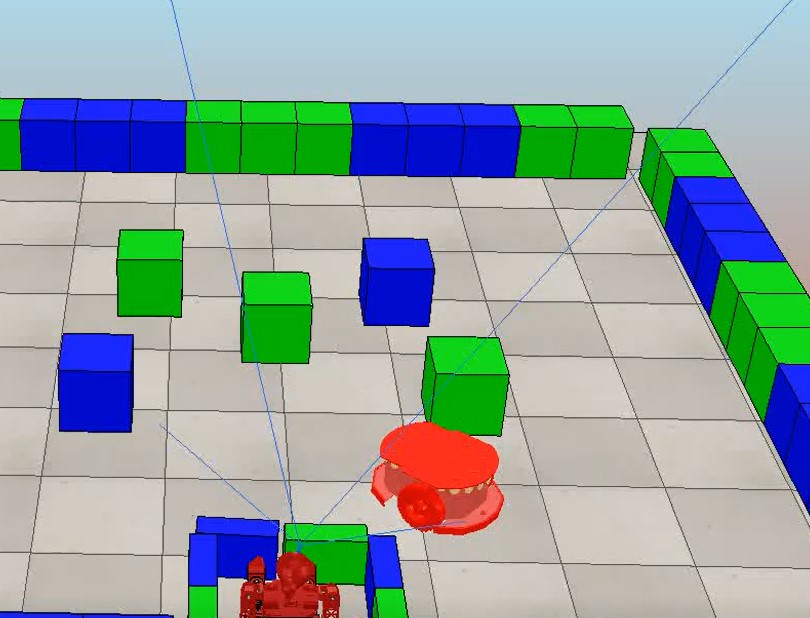}&
    \includegraphics[height=2.2 cm]{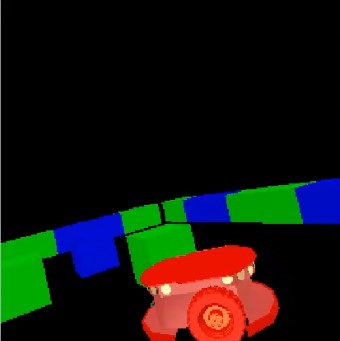}&
     \includegraphics[height=2.3cm]{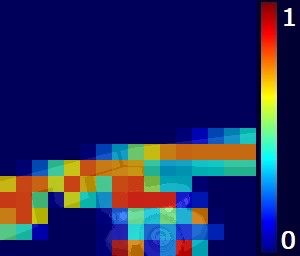}&
    \includegraphics[height=2.3cm]{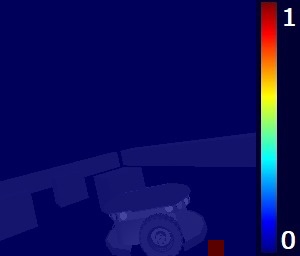}&
    \includegraphics[height=2.5cm]{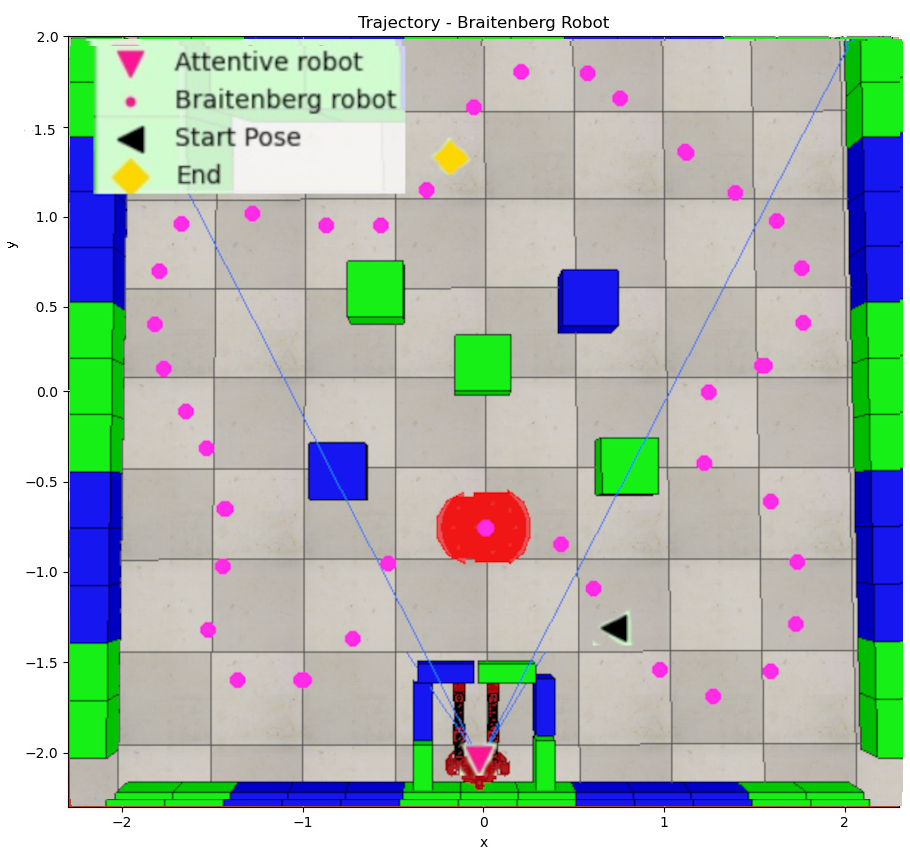} \\ \\
    
%    (a) Cena & (b) Sensor de visão & (c) Mapa de saliência &  (d) Vencedor \\

    \end{tabular}
    \end{adjustbox}
    \caption{2nd Substage. Sensory data obtained in the 1st episode of Procedural Learning. Left to right.
     (a) Overview of the scene in the simulator \emph{CoppeliaSim} (t = 1s);
     (b) Marta's camera view (t = 1s);
     (c) Salience Map (t = 3s);
     (d) Winner of the attentional cycle  (t = 3s);
     (e) Agents and objects' positions in scene.}
    \label{fig:exps_st2_att}

\end{figure*}

In this experiment, Marta continues with only \emph{bottom-up} perception elements. With the implementation of a motivation model, the agent starts to explore possible actions that do not have defined schemes in the Procedural Memory ${M}_{p}$. The reflex reactions developed in the 1st substage can generate primary circular reactions, stabilizing the learning of certain actions. The results obtained for Procedural Learning (training) in this experiment are shown in Figure \ref{fig:exps_st2_att}.
Using the QTable from the previous substage allows the agent to perform better in the object-tracking task.
The Attentional System of the 2nd Substage, during Procedural Learning, allowed the agent to establish its attentional focus on the Pioneer P3DX robot while it moved in regions closer to the humanoid, as in the 1st substage. With the withdrawal of the Pioneer P3DX robot, the agent returned to direct its attention to nearby objects and its own body. However, with the action of the motivation system, the agent was motivated to explore all possible actions for each new scheme not found in the Procedural Memory. This behavior minimized the performance of the reinforced actions developed in the 1st substage, allowing the agent to acquire greater rewards and promoting the formation of primary circular reactions.

\noindent \textbf{Learning Validation - 2nd Substage} 
\subsubsection{\textbf{Experiment A - 2nd Substage - Moving object with primary color}} 
\begin{figure*}[!h]
    \centering
    \begin{adjustbox}{max width=\textwidth}
     \begin{tabular}{ccccc}
    \includegraphics[height=2.2 cm]{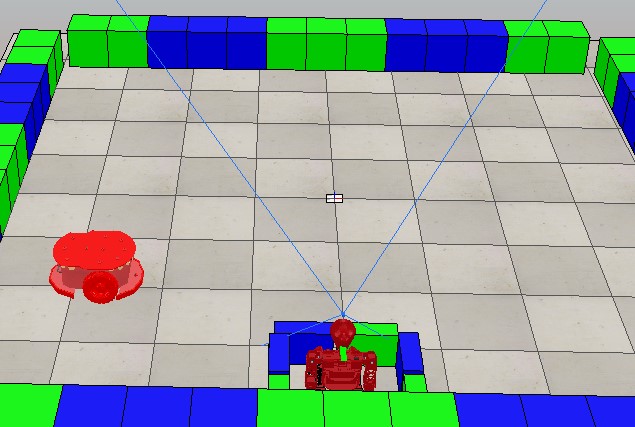}&
    \includegraphics[height=2.3 cm]{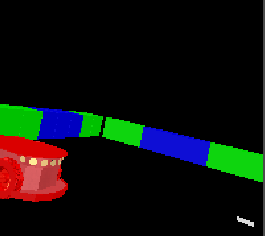}&
     \includegraphics[height=2.3cm]{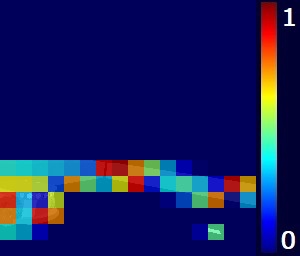}&
    \includegraphics[height=2.3cm]{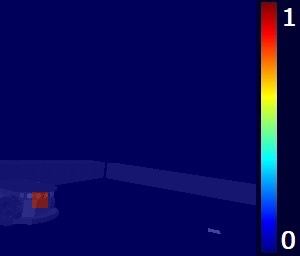}&
    \includegraphics[height=2.5 cm]{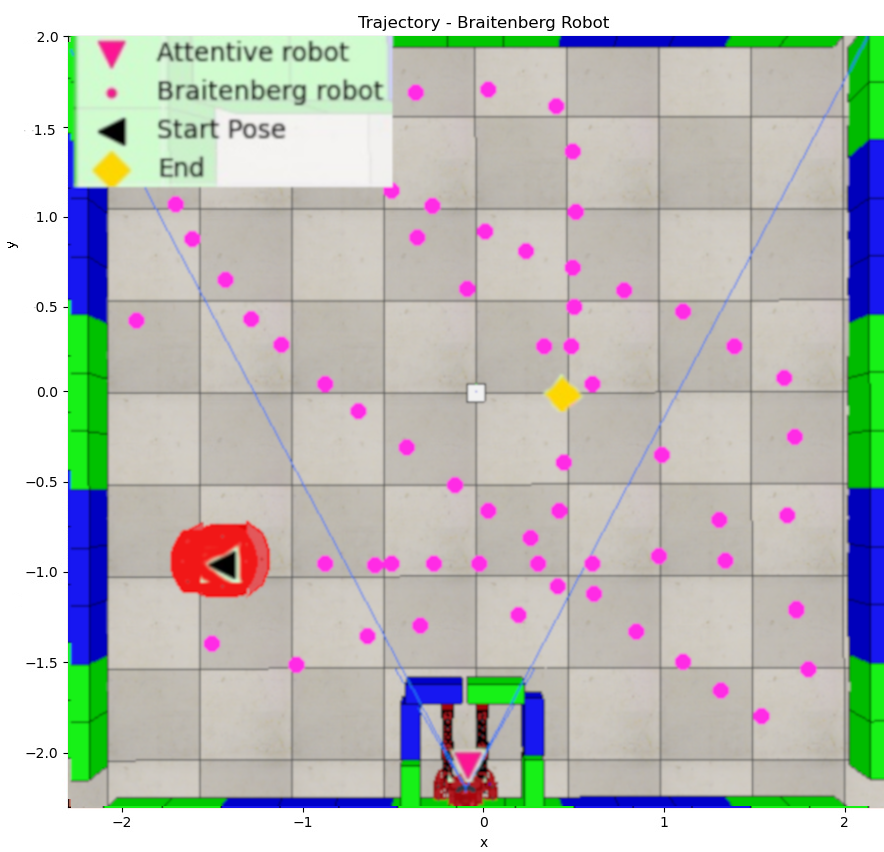}\\ \\
    
%    (a) Cena & (b) Mapa de saliência &  (c) Vencedor \\

    \end{tabular}
    \end{adjustbox}
    \caption{2nd Substage. Sensory data obtained in Experiment A. Left to right. (a) Overview of the scene in the simulator \emph{CoppeliaSim} (t = 1s);
    (b) Marta's camera view (t = 1s);
     (c) Salience Map (t = 3s);
     (d) Winner of attentional cycle (t = 3s);
     (e) Agents and objects' positions in scene.}
    \label{fig:exp_berto_sub2_a}

\end{figure*}

The QTable resulting from the end of the last episode of Procedural Learning was used. Marta was positioned with the Pioneer P3DX robot out of its field of view, as illustrated in the scene in Figure \ref{fig:exp_berto_sub2_a} (a). The Pioneer P3DX robot has a Braitenberg algorithm and moves at a constant speed of 0.1m/s. The agent directed its attention to the closest regions of its body when the Pioneer P3DX robot left its field of vision. The use of primary circular reactions promoted a performance with more actions performed compared to the previous substage. The increased visual acuity and the refinement of actuator movement in this substage also gave the agent greater control over its actuators.

\subsection{3rd Substage: Secondary Circular Reactions.} 
\begin{figure*}[!h]
    \centering
    \begin{adjustbox}{max width=\textwidth}
     \begin{tabular}{ccccc}
    \includegraphics[height=2.2 cm]{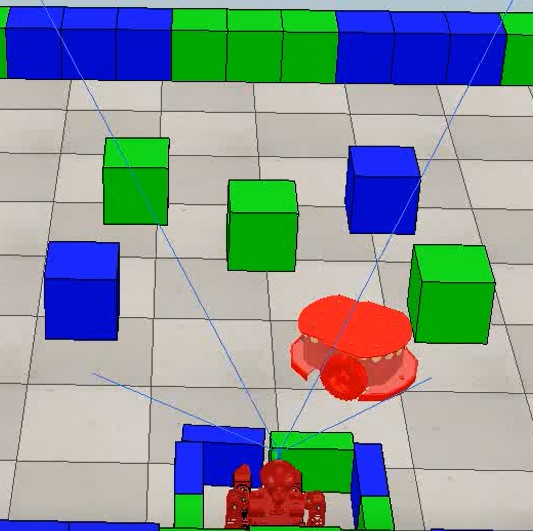}&
    \includegraphics[height=2.2 cm]{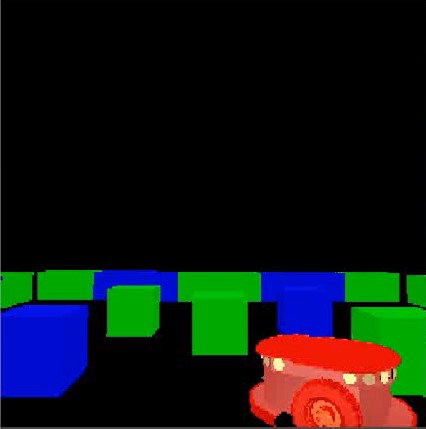}&
     \includegraphics[height=2.3cm]{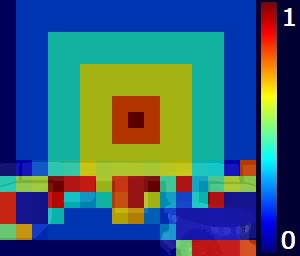}&
    \includegraphics[height=2.3cm]{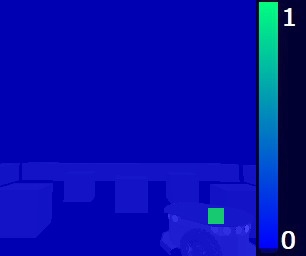}&
    \includegraphics[height=2.3cm]{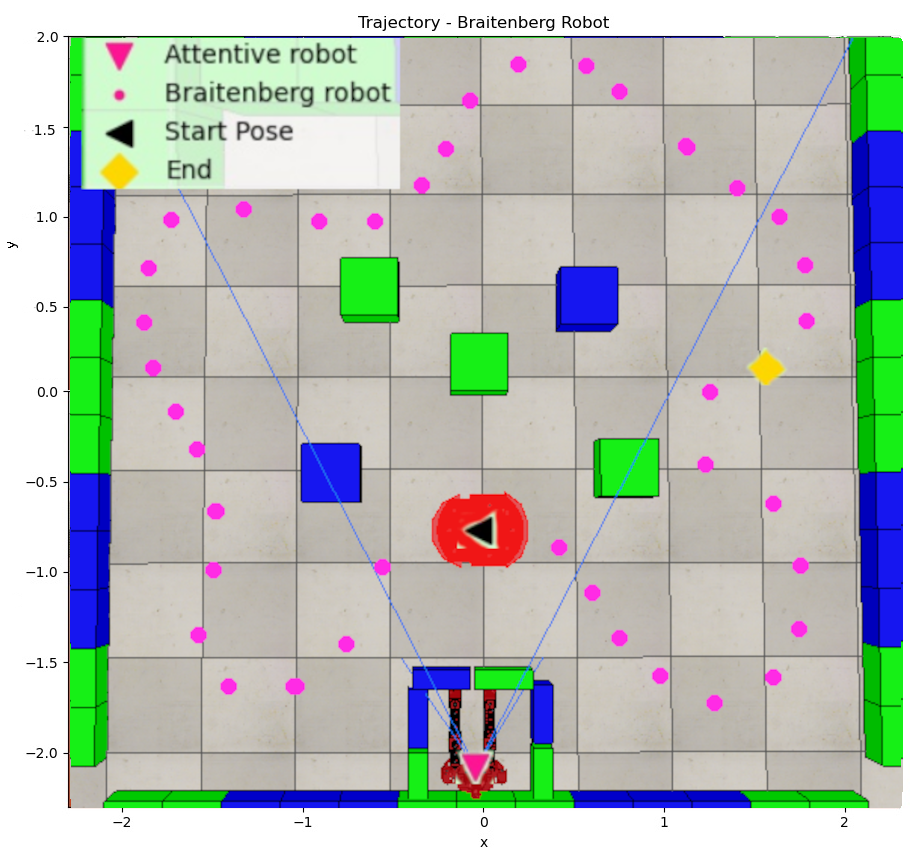}\\ \\
    
%    (a) Cena & (b) Sensor de visão & (c) Mapa de saliência &  (d) Vencedor \\

    \end{tabular}
    \end{adjustbox}
    \caption{ 3rd Substage. Sensory data obtained in the 1st episode of Procedural Learning. Left to right. (a) Overview of the scene in the simulator \emph{CoppeliaSim} (t = 1s);
     (b) Marta's camera view (t = 1s);
     (c) Salience Map (t = 3s);
     (d) Winner of attentional cycle  (t = 3s);
     (e) Agents and objects' positions in scene.}
    \label{fig:exps_st3_att}
  \end{figure*}

In this substage, the humanoid Marta has a cognitive-attentional algorithm that has all the elements shown in Figure \ref{fig:model}, with elements of perception \emph{bottom-up} and \emph{top-down}, and models of intentionality and motivation. The agent can exploit the primary circular reactions developed in the previous steps and develop secondary circular reactions. The results obtained for Procedural Learning (training) in this experiment are shown in Figure \ref{fig:exps_st3_att}.
The Attention System of the 3rd Substage, during Procedural Learning, allowed the agent to establish its focus of attention on the P3DX robot. By using attentional actions, the agent could follow the movement of the P3DX even when it was in the most distant regions of the humanoid and out of its field of vision.

  \noindent \textbf{Learning Validation - 3rd Substage} 
\subsubsection{\textbf{Experiment A - 3rd Substage - Moving object and primary color.}} The QTable resulting from the last episode of Procedural Learning was used. We positioned Marta with the Pioneer P3DX robot out of its field of view. The Pioneer P3DX robot uses the Braitenberg algorithm and moves at a constant speed of 0.1m/s. In this experiment, the agent maintained its focus on the Pioneer P3DX robot even when it was far away because the agent was able to track the moving robot. This is mostly caused by motivation and intention to follow the moving object. The increased visual acuity in this substage and the refinement of actuator movement gave the agent greater control over its actuators than in previous substages. The secondary circular reactions developed during this substage promoted a higher performance than in previous substages, as we can see in Figure \ref{fig:results_actions}.

\begin{figure*}[!h]
    \centering
    \begin{adjustbox}{max width=\textwidth}
     \begin{tabular}{ccccc}
    \includegraphics[height=2.2 cm]{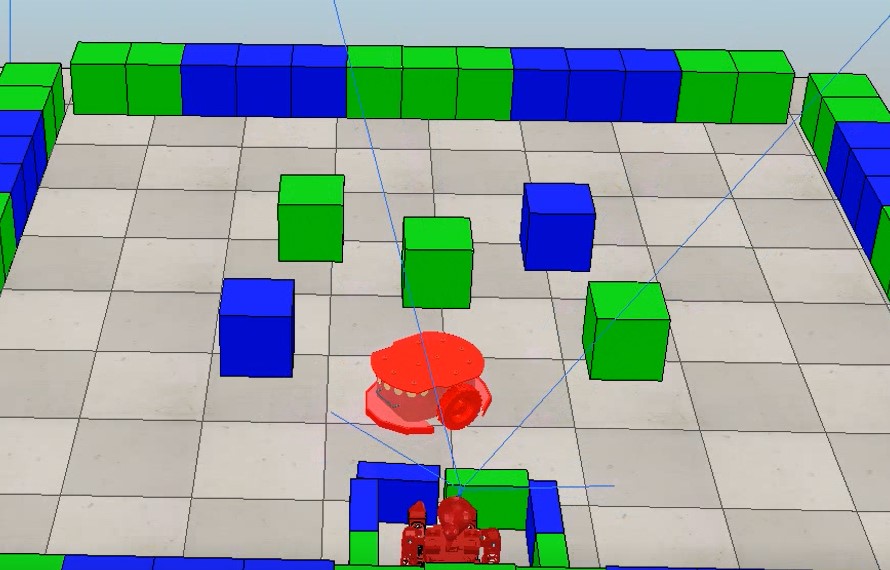}&
    \includegraphics[height=2.2 cm]{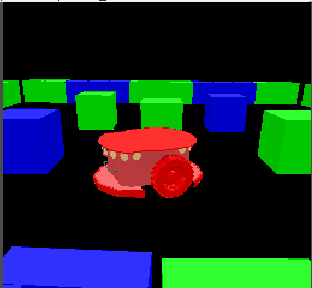}&    
     \includegraphics[height=2.3cm]{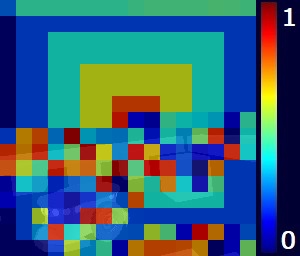}&
    \includegraphics[height=2.3cm]{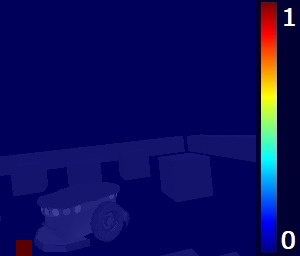}& 
    \includegraphics[height=2.5cm]{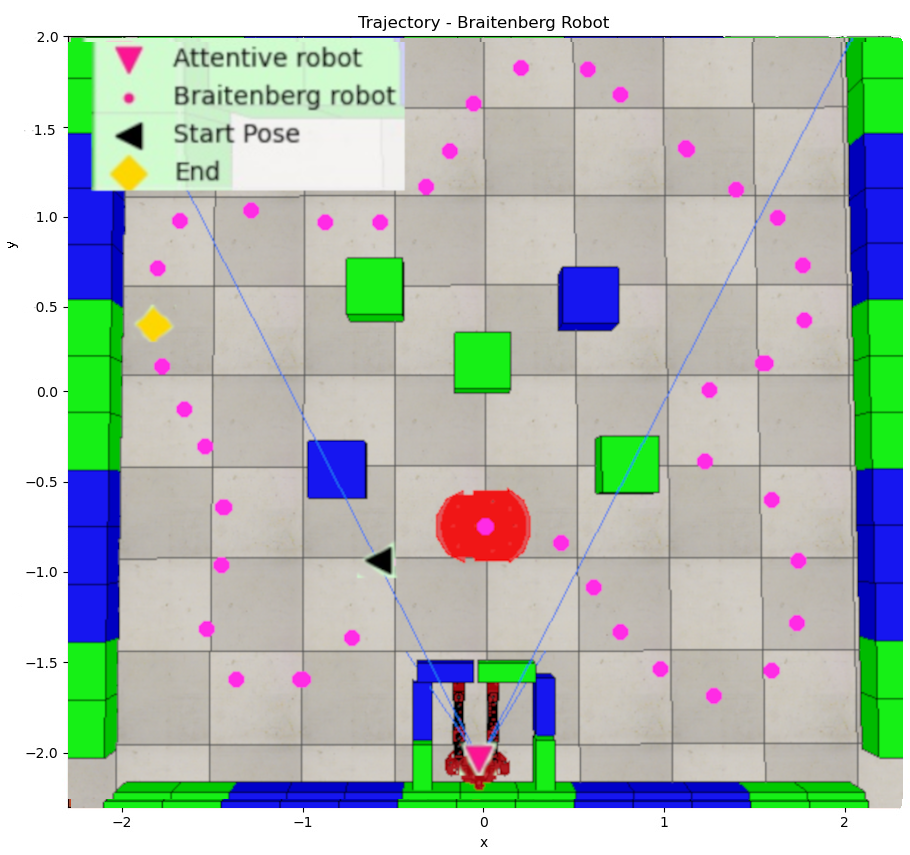}\\ \\
    
%    (a) Cena & (b) Mapa de saliência &  (c) Vencedor \\

    \end{tabular}
    \end{adjustbox}
    \caption{3rd Substage. Sensory data obtained in validation experiment A. Left to right. (a) Overview of the scene in the simulator \emph{CoppeliaSim} (t = 1s);
    (b) Marta's camera view (t = 1s);
     (c) Salience Map (t = 3s);
     (d) Winner of the attentional cycle  (t = 3s);
     (e) Agents and objects' positions in scene.}
    \label{fig:exp_berto_sub3_a}

\end{figure*}

\section{Conclusion}
In this work, we proposed and implemented an incremental procedural learning mechanism to create and reuse previously learned schemas inspired by Piaget's sensorimotor development substages. To this end, we employed a simulated humanoid robot and static and moving objects. By building our cognitive agent, we investigated which modules in a cognitive architecture are needed to control a robot interacting with its environment while performing a set of sensorimotor experiments with increasing difficulty. We discussed the importance of motivation and attention to forming primary and secondary circular reactions from reflexes. This approach allowed for employing a single incremental mechanism that evolves over time. We showed that reusing previous knowledge is mandatory for the success of incremental learning. The experiments demonstrated the feasibility of using a cognitive-attentional architecture based on CONAIM and implemented with CST. Furthermore, we successfully implemented experiments corresponding to the first three substages proposed by Berto (2020) \cite{berto_2020_thesis} for tracking objects. With these experiments, we could show which cognitive functions are required to achieve specific levels of development through object-tracking experiments.

\section{Acknowledgments}
\noindent This work was developed within the scope of PPI-Softex with support from MCTI through the Technical Cooperation Term [01245.013778/2020-21].

\bibliographystyle{IEEEtran}

\bibliography{IEEEtran}

\begin{IEEEbiography}[{\includegraphics[width=1in,height=1.25in,clip,keepaspectratio]{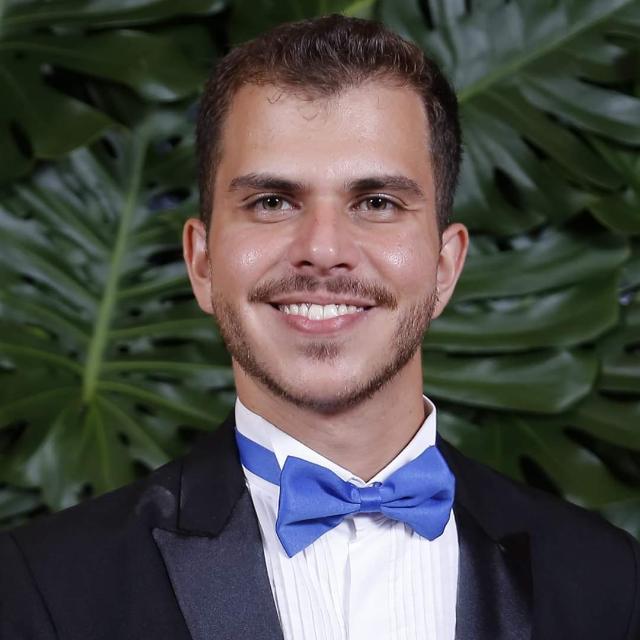}}]{Leonardo de Lellis Rossi} graduated in Control and Automation Engineering and received a Master's degree in Electrical Engineering from the Institute of Science and Technology of Sorocaba (ICTS) - Unesp. He is a Ph.D. candidate in Electrical Engineering at the Faculty of Electrical and Computing Engineering (FEEC/UNICAMP) and a member of the Cognitive Architectures area at H.IAAC (Hub of Artificial Intelligence and Cognitive Architectures). His research interests include Cognitive Robotics, Developmental Robotics, Machine Learning, and Deep Learning.\end{IEEEbiography}

\begin{IEEEbiography}[{\includegraphics[width=1in,height=1.25in,clip,keepaspectratio]{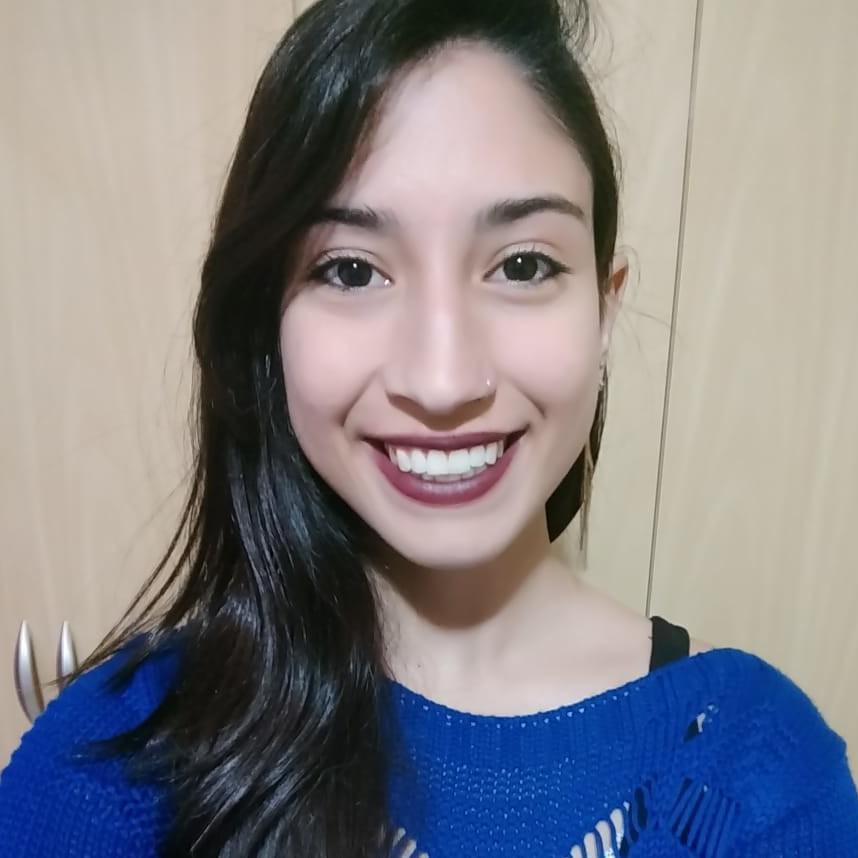}}]{Letícia Mara Berto} has a B.S. degree in Computer Science at Federal University of São Carlos (UFSCar), Brazil, an M.S. degree in Computer Science at the Institute of Computing (IC), State University of Campinas (Unicamp), Brazil. She is a Ph.D. candidate at the Institute of Computing (IC), State University of Campinas (Unicamp), Brazil, and is visiting researcher at the Italian Institute of Technology (IIT), Genova, Italy. Her research interests include cognitive robotics, affective computing and human-robot interaction. 
\end{IEEEbiography}

\begin{IEEEbiography}[{\includegraphics[width=1in,height=1.25in,clip,keepaspectratio]{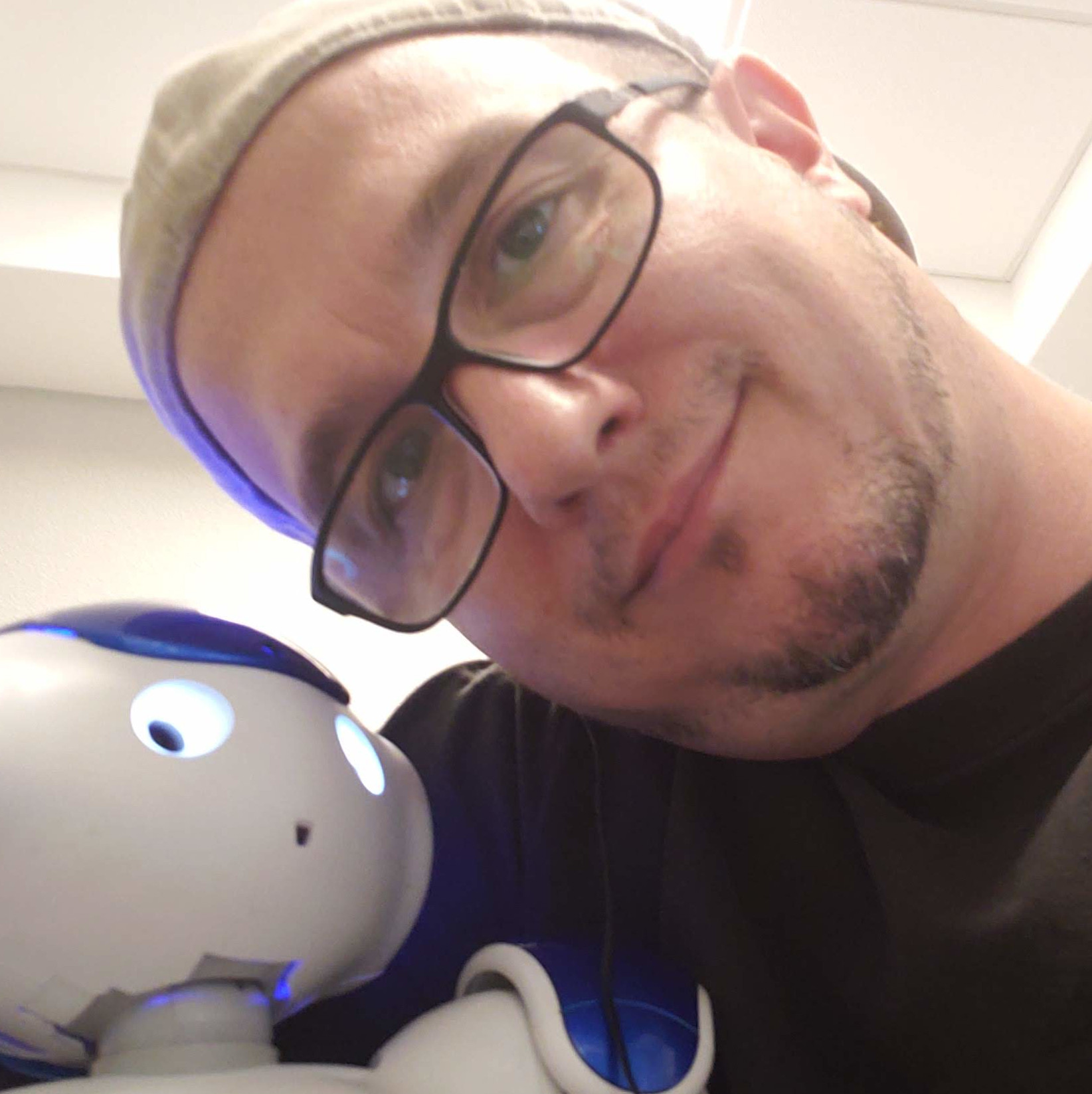}}]{Eric Rohmer} is a Professor at the Faculty of Electrical and Computer Engineering, State University of Campinas (Unicamp), Brazil. He received the M.S. in 2000 at the \textit{Ecole Supérieure d'Informatique et Applications de Lorraine}, Nancy, France, and the Ph.D. in Computer Science in 2005 at the Advance Robotics Laboratory, Sendai, Japan. He acted as a researcher at the Space Robotic Laboratory and at the Rescue Robotics Laboratory in Japan. His research interests include planetary exploration, rescue robots, mobility on rough terrain, reconfigurable mobile robots, assistive robotics and robot control.  \end{IEEEbiography}

\begin{IEEEbiography}[{\includegraphics[width=1in,height=1.25in,clip,keepaspectratio]{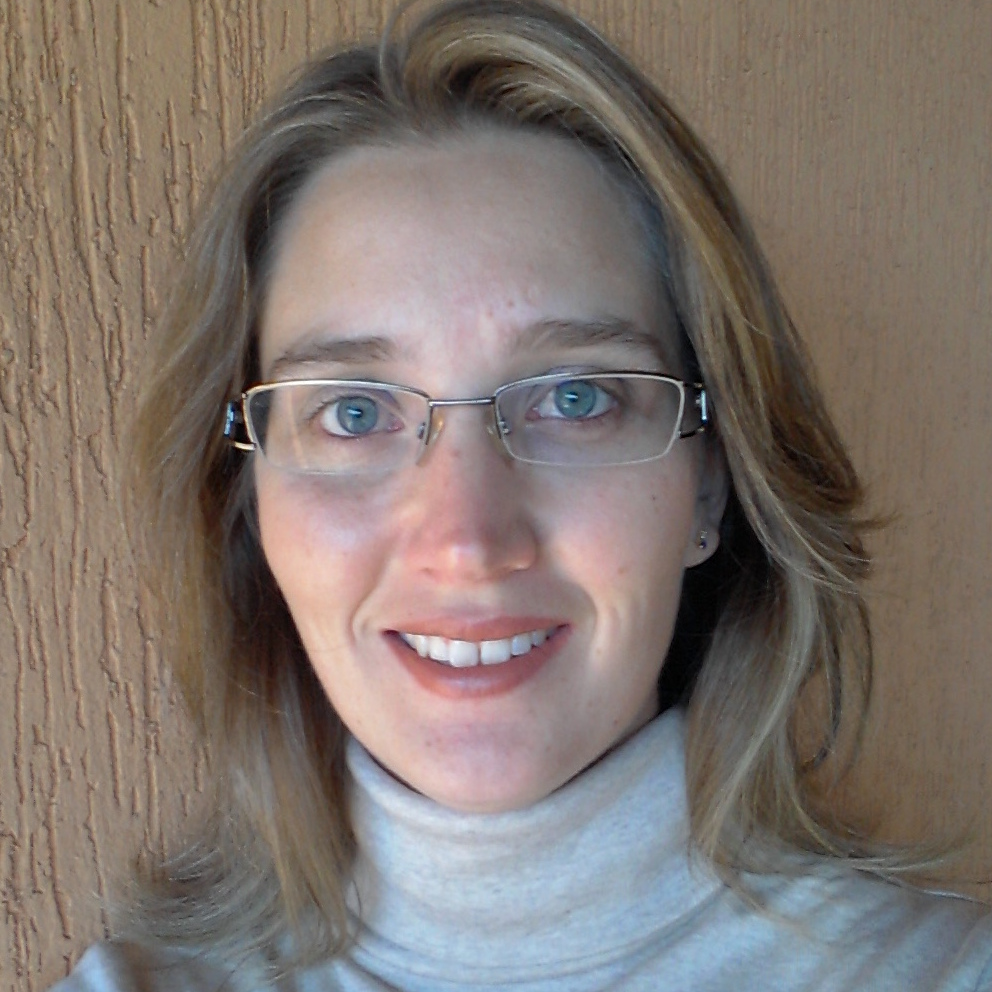}}]{Paula Paro Costa} is a Professor at the Faculty of Electrical and Computer Engineering, State University of Campinas (Unicamp), Brazil. She received a B.S. degree in Electrical Engineering in 2001 at the State University of Campinas, Brazil, an M.S. in Mobile Communication in 2006 at the Polytechnique of Torino, Italy, an M.S. (2009) and the Ph.D. (2015) in Computer Engineering at State University Of Campinas, Brazil. Her research interests include multimodal artificial intelligence, affective computing, and science disclosure for young people. \end{IEEEbiography}

\begin{IEEEbiography}[{\includegraphics[width=1in,height=1.25in,clip,keepaspectratio]{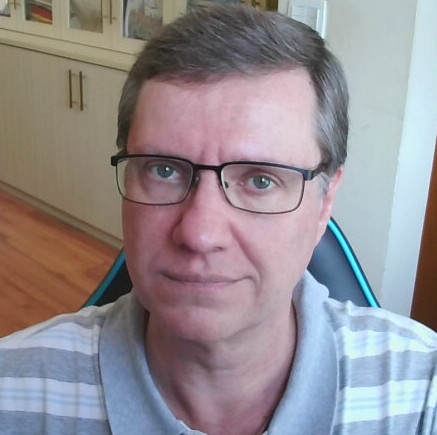}}]{Ricardo Ribeiro Gudwin} is an Associate Professor at the Faculty of Electrical and Computer Engineering, State University of Campinas - Brazil. He was born in Campinas-SP, Brazil, in 1967. He received a B.S. degree in Electrical Engineering in 1989, an M.S. degree in Electrical Engineering in 1992, and a Ph.D. in Electrical Engineering in 1996, all from the Faculty of Electrical and Computer Engineering, State University of Campinas - Brazil. His earlier research interests include fuzzy, neural, and evolutionary systems. His current research interests include the study of cognitive architectures, cognitive systems, intelligence and intelligent systems, intelligent agents, semiotics, and computational semiotics. \end{IEEEbiography}

\begin{IEEEbiography}[{\includegraphics[width=1in,height=1.25in,clip,keepaspectratio]{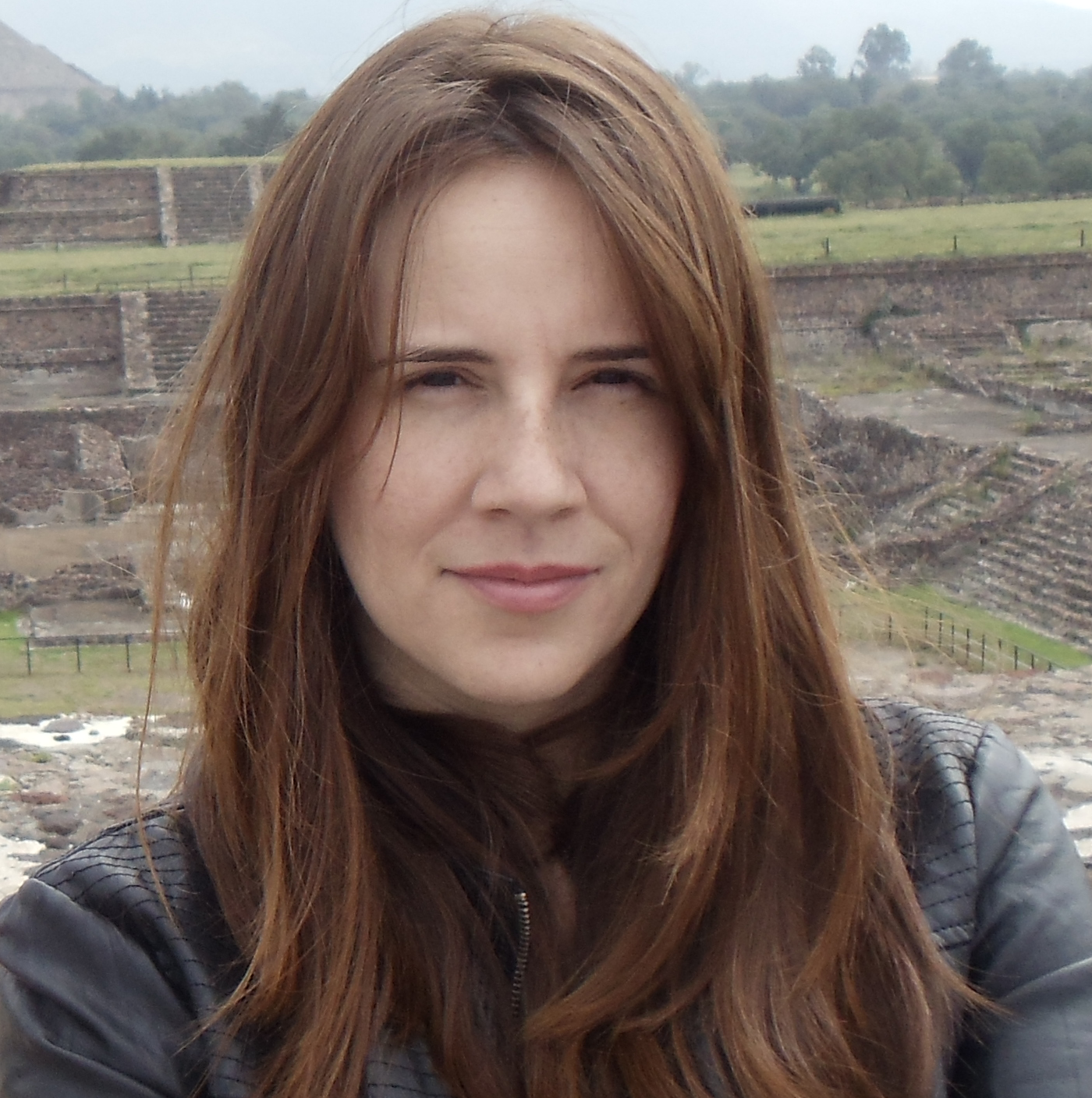}}]{Esther Luna Colombini} is a Professor at the University of Campinas, Brazil, where she coordinates the Laboratory of Robotics and Cognitive Systems (LaRoCS). She holds an M.Sc. and a Ph.D. in computing engineering from the Technological Institute of Aeronautics, Brazil. Her research interests include machine learning and AI for autonomous robotics and healthcare, focusing on Reinforcement Learning, Attentive Models and Cognitive Modeling. She is a research member of the Advanced Institute for Artificial Intelligence (AI2), the Brazilian Institute of Neuroscience and Neurotechnology (BRAINN), and the Center for Artificial Intelligence (C4Ai). She acted as President of RoboCup Brazil, co-founder, and chair of the Brazilian Robotics Olympiad and the National Robotics Fair, two public initiatives - supported by the National Council of Technological and Scientific Development - reaching yearly over 230,000 students aiming at attracting them to STEM careers. She coordinates the multilateral BRICSmart Alliance supported by CNPq and the BRICS-STI committee to employ ICT for Smart Resource Utilization to Combat Global Pandemic Outbreaks and the Learning in Cognitive Architectures research in the Hub for Artificial Intelligence and Cognitive Architectures (H.IAAC).  \end{IEEEbiography}

\begin{IEEEbiography}[{\includegraphics[width=1in,height=1.25in,clip,keepaspectratio]{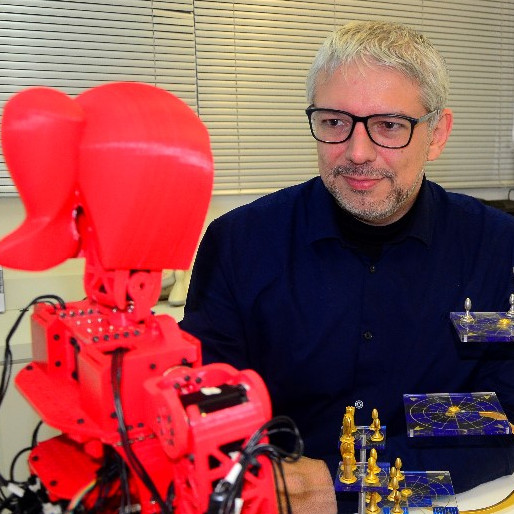}}]{Alexandre da Silva Simões} is an Associate Professor at the Control and Automation Engineering Department (DECA) at Institute of Science and Technology (ICT), Campus Sorocaba, São Paulo State University (Unesp), Brazil. He received a B.S. degree in Electrical Engineering at Unesp, Campus Bauru, in 1998, an M.S. degree in Electrical Engineering (2000), and a Ph.D. in Electrical Engineering (2006), both from the Polytechnic School of the São Paulo University (USP), Brazil. His research interests include robotics, machine learning, cognitive systems, and the interface between these research fields, science disclosure, education, and arts. \end{IEEEbiography}

\end{document}